\documentclass[twoside,11pt]{article}

\usepackage[T1]{fontenc}

\usepackage{enumitem}

\usepackage[preprint]{jmlr2e}

\usepackage[ruled,vlined,linesnumbered]{algorithm2e}

\usepackage{hyperref}
\usepackage{amsmath}
\usepackage{longtable}
\usepackage{pdflscape}
\usepackage[commandnameprefix=ifneeded]{changes}

\usepackage{tikz}

\usepackage{natbib}
 \bibpunct[, ]{(}{)}{,}{a}{}{,}%
\begin{document}

\title{Active Learning with Imperfect Labels: Optimal Labeler Assignment and Sample Selection}

\author{
    \name Pouya Ahadi \email \href{mailto:pouya.ahadi@gatech.edu}{pouya.ahadi@gatech.edu} \\
    \addr H. Milton Stewart School of Industrial and Systems Engineering\\
    Georgia Institute of Technology\\
    Atlanta, GA 30332, USA
    \AND
    \name Blair Winograd \email \href{mailto:bawinogr@gmail.com}{bawinogr@gmail.com} \\
    \addr Ford Motor Company
    \AND
    \name Camille Zaug \email \href{mailto:czaug@ford.com}{czaug@ford.com} \\
    \addr Ford Motor Company
    \AND
    \name Karunesh Arora \email \href{mailto:karora4@ford.com}{karora4@ford.com} \\
    \addr Ford Motor Company
    \AND
    \name Lijun Wang \email \href{mailto:lwang149@ford.com}{lwang149@ford.com} \\
    \addr Ford Motor Company
    \AND
    \name Kamran Paynabar \email \href{mailto:kamran.paynabar@isye.gatech.edu}{kamran.paynabar@isye.gatech.edu} \\
    \addr H. Milton Stewart School of Industrial and Systems Engineering\\
    Georgia Institute of Technology\\
    Atlanta, GA 30332, USA
}

\maketitle

\begin{abstract}
Active Learning (AL) is commonly used in applications where labeling data is expensive or time-consuming. In practice, however, labels are often noisy due to varying labeler expertise and annotation uncertainty, especially for complex or ambiguous samples. Learning from such imperfectly labeled data can degrade classifier performance. We propose an AL framework that explicitly accounts for label noise by optimally assigning labelers and selecting samples to minimize labeling error. Our approach, called OLAS (Optimal Labeler Assignment and Sampling), uses a noise model that depends on both labeler accuracy and model uncertainty to guide these decisions. We develop two tractable optimization formulations: one for assigning samples to labelers to minimize worst-case noise, and another for selecting samples while controlling overall label noise. Theoretical results provide closed-form solutions under mild conditions. {Empirical evaluations on benchmark datasets and a real-world warranty claim classification problem show that OLAS achieves the highest or near-highest classification accuracy among existing AL strategies across most settings, using only a single label per sample.}
\end{abstract}




\maketitle
\section{Introduction}
\label{sec:intro}

In typical classification problems, machine learning models are trained using a set of labeled data points, with these labels being assigned through annotation. In numerous applications, the annotation of samples can be a time-consuming or costly endeavor. For instance,~\cite{singh2009active} explored the challenges of image classification, where annotating images can be labor-intensive. Similarly, in the field of additive manufacturing, acquiring labels has become increasingly time- and labor-intensive, especially with the integration of advanced sensors that generate high-dimensional, large datasets~\citep{van2021active}. Given the vast volumes of datasets in many such applications, fully annotating all instances has become an exceedingly challenging, if not impossible, task. As a result, the training of these models often relies on a limited set of labeled data.

The limitation in annotating unlabeled samples naturally establishes a labeling budget defined as the number of unlabeled observations that can be annotated. Consequently, selecting a subset for annotation becomes a critical task. In general learning problems, there are typically two strategies for choosing the best samples: \emph{one-shot sampling} methods and \emph{sequential sampling} methods. In the one-shot sampling approach, a fixed number of observations are chosen based on the available budget. Once the labels for all these samples are obtained, a model can be trained using this labeled training data \citep{johnson1990minimax, santner2018space}. Conversely, in the \emph{sequential sampling} design, new observations are selected over time, informed by data from previously collected samples. At each step, these accumulated samples are used to update the training model, and the information gained from the most recently updated model guides the selection of subsequent samples~\cite{gahrooei2019adaptive, huang2023proportional}. Therefore, while one-shot sampling methods do not utilize information from earlier observations, sequential sampling methods effectively address this limitation by iteratively updating the model with the most recent observations.

By employing sequential sampling, the total labeling budget can be distributed across several cycles. In each cycle, a number of observations are sampled and labeled, based on the budget allocated for that cycle. Subsequently, the model is updated with this new batch of labels. In the context of Machine Learning, this procedure is termed \emph{active learning (AL)}. In AL, the key challenge is selecting samples in each cycle by leveraging insights from the most recently updated model. A common approach known as \emph{uncertainty sampling} assumes that observations with higher uncertainty offer more information about the underlying class distribution, making them ideal candidates for labeling~\citep{settles2012active}. There are various methods to quantify uncertainty, leading to different uncertainty sampling strategies, including \emph{least confident}~\citep{settles2008analysis}, \emph{margin sampling}~\citep{scheffer2001active}, and \emph{entropy sampling}~\citep{li2019entropy}. Among these, entropy sampling, which employs the entropy measure for sampling the observations, is arguably the most widely used method in AL. Therefore, our focus in this paper will be on entropy sampling.

Annotating data points involves assigning labels to unlabeled samples, typically performed by various labelers (oracles). However, in practical scenarios, these labels can often be incorrect due to noisy labelers as noted by~\cite{gupta2019learning}. As a result, the labels obtained might be erroneous, and training a model with such incorrect labels can lead to an inaccurate prediction model.~\cite{abdellatif2021active} investigated the issue in the context of connected vehicles, where a single road event may be classified into various classes using multiple vehicles simultaneously. Each vehicle is equipped with a weak prediction model which can generate noisy labels. A critical aspect of noisy labeling is that observations with higher uncertainty are more prone to label noise, as discussed by~\cite{du2010active}. This suggests that samples for which the classifier is less confident are also likely to be less certain for the labelers. This presents a challenge in uncertainty sampling methods, as the tendency to select uncertain observations can unintentionally increase susceptibility to label noise.

Another motivating application of the problem described above is \emph{automated claim management (ACM)} for warranty claims in manufacturing. {Companies receive warranty claims on a daily basis, and each claim must be categorized into a predefined category.} ACM employs machine learning models to categorize all incoming warranty claims. Concurrently, there is a manual process of binning these claims, performed by technicians who act as labelers and provide data for updating the classification model. Due to the large volume of warranty claim data, technicians can only label a subset of the claims. This limitation motivates the integration of Active Learning (AL) into the claim-binning process, leading to the development of an AL-based ACM process for the company. In each AL cycle, selected observations are sent to technicians for labeling, and the received labels are then used to update the classification model. However, due to differences in skill and experience levels, technicians may make mistakes during the manual binning process, resulting in noisy labels. This motivates the need for a systematic approach to handle noisy labelers in active learning settings.

The main goal of this paper is to propose a novel active learning (AL) framework for selecting informative samples and assigning them to labelers, considering their skill levels and labeling noise. We refer to our method as \emph{OLAS} (Optimal Labeler Assignment and Sampling for Active Learning).
 We examine the process of noise generation within the AL framework and develop a formulation for assigning query points to labelers, aiming to minimize worst-case noise. Additionally, we adapt the entropy sampling approach to address the challenge of imperfect labels by reformulating it as an optimization model. We also introduce an efficient method for solving this model to optimality.

{
Much prior work addresses noisy annotations from multiple labelers by aggregating or reconciling labels after they have been collected. The classical Dawid and Skene model \citep{dawid1979maximum} estimates true labels and labeler reliabilities jointly via expectation maximization. \citet{rodrigues2013learning} propose a related probabilistic model that treats each annotator's reliability as a latent Bernoulli variable, reducing the number of parameters relative to full confusion matrix approaches and thereby avoiding overfitting when each annotator labels only a small fraction of the data, a setting we return to directly in Section~\ref{sec.music} using the same music genre classification data introduced in that paper. In a similar spirit, \citet{sheng2008get} study how to improve data quality by requesting repeated labels for the same instance from multiple noisy labelers, and \citet{zhao2011incremental} propose incremental relabeling strategies that identify and re-annotate instances whose labels appear inconsistent across the crowd. \citet{kaluza2023robust} propose annotation unification algorithms that estimate annotator reliability and reconcile labels from sparse, largely non-overlapping annotators, requiring little to no intersection between the samples seen by different labelers. All of these approaches share a common structure: labels are collected, often redundantly or from overlapping annotators, and quality is improved by aggregating or reconciling them after acquisition.}

{Other work brings this same correction after acquisition into the active learning loop itself, rather than treating it as a separate offline step. \citet{wang2023active} allocate part of the labeling budget to relabeling instances suspected of being mislabeled, rather than exclusively acquiring labels for new instances, and \citet{younesian2021qactor} similarly filter already-labeled instances into clean and suspected-noisy groups and send the most informative suspected-noisy instances back to an oracle for relabeling. Both approaches improve label quality by revisiting samples after an initial, possibly incorrect, label has already been obtained. By contrast, \citet{zhang2015active} intervene earlier, before any label is collected for a given instance, by studying active learning with both weak and strong labelers and addressing the trade off between labeler quality and cost; to our knowledge, this is the closest existing precedent for deciding, proactively, which labeler should annotate a previously unlabeled instance, though their method selects which instances to query under label noise without jointly optimizing which specific labeler receives each instance as part of the same decision.}

{A closer line of work performs sample and labeler selection jointly. \citet{huang2017cost} propose an active selection criterion that evaluates instance-labeler pairs by cost-effectiveness, exploiting the fact that a labeler with lower overall quality can still be accurate on instances within their specific expertise. \citet{fang2014active} transfer knowledge from unlabeled auxiliary domains to estimate labeler expertise, then jointly select the most informative instance and the labeler with the best expertise for it. \citet{zhao2014active} jointly estimate worker performance, task difficulty, and annotation reliability within a Bayesian network, using the resulting risk estimates to guide sample selection. \citet{long2013active} estimate each labeler's expertise alongside a Gaussian process classifier and actively select both which instance to query and which labeler should annotate it. OLAS differs from these four along several dimensions: all four permit repeated or redundant labeling across labelers, whereas OLAS restricts each sample to a single label; none models an explicit per-cycle labeler capacity constraint, whereas OLAS does; each selects a labeler to maximize expected label quality or minimize expected risk for the queried instance, whereas OLAS minimizes the maximum noise across all assigned labels, a worst-case rather than average-case criterion; and each relies on heuristic selection criteria or approximate inference, whereas OLAS admits closed-form optimal solutions (Theorems~\ref{thm.mm} and~\ref{thm.qm}) and a polynomial-time algorithm (Algorithm~\ref{alg}).}

{OLAS differs from this literature in how it handles noisy annotations. Unlike aggregation, reconciliation, and relabeling strategies such as Dawid and Skene \citep{dawid1979maximum}, \citet{rodrigues2013learning}, \citet{sheng2008get}, \citet{zhao2011incremental}, \citet{kaluza2023robust}, \citet{wang2023active}, and \citet{younesian2021qactor}, OLAS never collects more than one label per sample and never revisits or reconciles a sample's label once it has been acquired; there is no redundancy to aggregate and no correction step to design. This distinction matters whenever repeated, overlapping, or revisited annotation is not merely undesirable but infeasible, for example when each additional label carries meaningful cost or delay, as in our motivating warranty claim application in Section~\ref{sec.experiments}. Instead, OLAS uses its labeling budget to reduce the expected noise of the single acquired label for each instance, which requires deciding sample selection and labeler assignment together, before any label exists, rather than as a two-stage process of first labeling and then correcting.}

{Relative to \citet{huang2017cost}, \citet{fang2014active}, \citet{zhao2014active}, \citet{long2013active}, and \citet{zhang2015active}, the closest precedents for joint or proactive labeler assignment, OLAS additionally restricts each sample to a single label under explicit labeler capacity constraints, and optimizes a worst-case rather than average-case noise objective, as detailed above. OLAS formulates this proactive, joint decision, of which sample to query and which single labeler should annotate it, as one optimization problem, for which we derive closed-form optimal solutions (Theorems~\ref{thm.mm} and~\ref{thm.qm}) and a polynomial-time algorithm (Algorithm~\ref{alg}), rather than relying on heuristic or approximate assignment rules. By proactively assigning each sample to a single labeler based on expected noise, driven jointly by model uncertainty and labeler reliability, OLAS addresses the labeling challenge through a single optimization of acquisition and assignment, avoiding post-hoc corrections or aggregation and making it suitable for high-cost real-world settings where repeated labeling is infeasible.
}

{The remainder of the article is organized as follows. 
Section~\ref{sec.overview} provides an overview of the active learning framework 
and situates the proposed methodology within this framework. 
Section~\ref{sec.methodology} presents the proposed approach and formulates two 
optimization models for handling noisy labels. 
Section~\ref{sec.experiments} evaluates the proposed framework through numerical 
experiments on publicly available datasets as well as a case study on ACM data. 
Finally, Section~\ref{sec.conclusion} concludes the paper.
}


\section{Overview of the Framework} \label{sec.overview}

The AL framework starts with some initially labeled data points. A classifier is then trained with these initially labeled instances. Due to labeling limitations in an AL paradigm, we can only label a partial subset of the unlabeled samples. Hence, we train and update the model sequentially, based on a subset of newly labeled training samples in each cycle or iteration. Suppose there is a limited budget for labeling only $T \times B$ instances, where $B$ is the labeling budget in each cycle and $T$ is the number of cycles. We denote cycles by $t = 1, 2, \dots, T$. 

\begin{figure}[t]
    \centering
\includegraphics[scale =0.13]{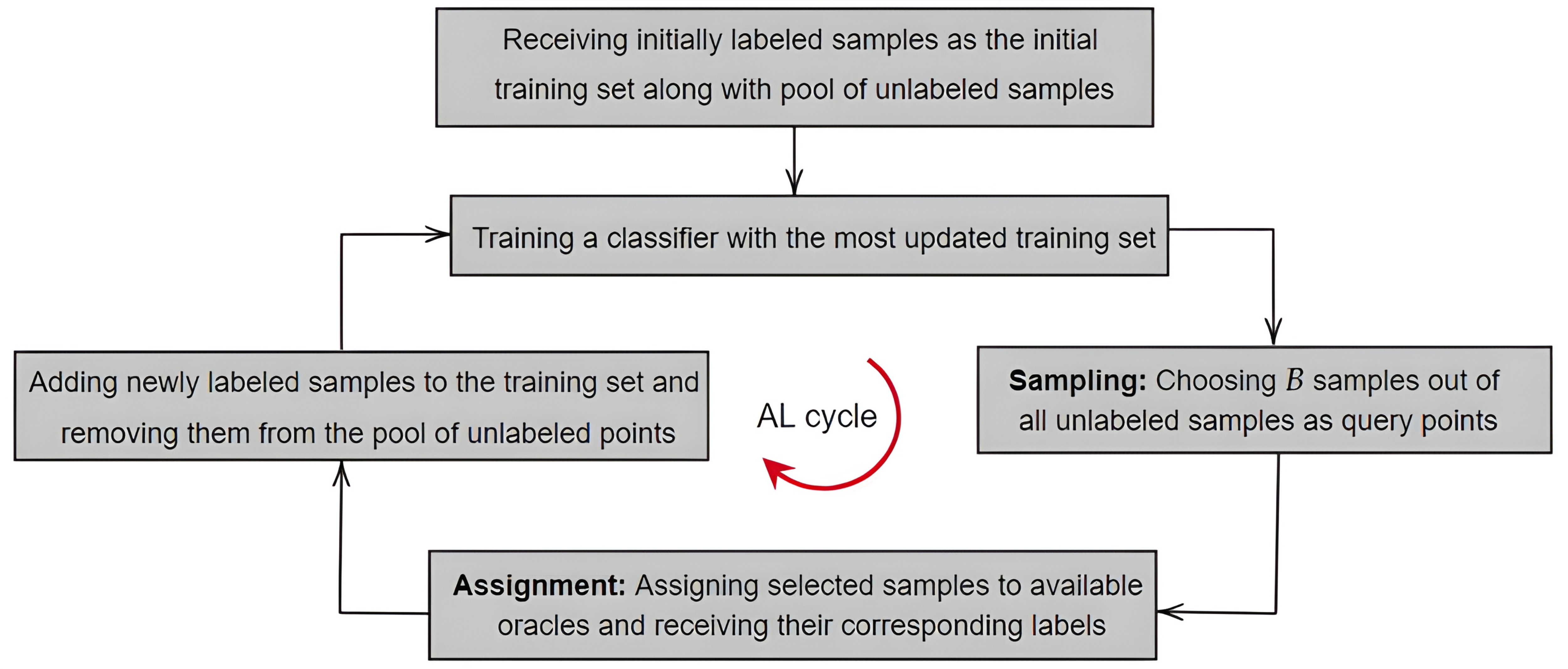}
    \vspace{0.1cm}
    \caption{General framework for AL for classification. The AL cycle will be repeated for $T$ cycles until the total sampling budget is consumed.}
    \label{fig.framework}
\end{figure}

At each cycle, a sampling criterion is utilized to choose $B$ samples from the pool of unlabeled instances. These selected samples are then assigned to labelers according to the optimization model we propose in the next section, and their labels are queried. The new labels are added to the training set and the model is updated based on the new training set. This process will be repeated for $T$ cycles. Figure~\ref{fig.framework} illustrates the overall AL framework. 

Due to noisy oracles/labelers, we may receive incorrect labels in each cycle, which can degrade the performance of the classifier over time. To maintain an effective active learning (AL) process, it is essential to account for labeling noise in each cycle, both during sampling and assignment. To address this, we model the label noise and apply optimization techniques to propose a sampling strategy, followed by an assignment policy that together form a robust AL framework. We will first study the assignment problem and then propose the optimal sampling approach based on the findings from the labeler assignment problem.


\section{Methodology} \label{sec.methodology}
An AL problem starts with the query selection where the set of unlabeled instances will be investigated to select a set of samples called \emph{the query set.} The selection of the query set depends on the sampling strategy. For now, assume that we have used a specific sampling strategy and obtained a query set $Q_t \subseteq U_t$, where $Q_t$ and $U_t$ denote the selected query set and set of unlabeled data points at Cycle $t$ of AL. 

After selecting the query set, each sample in the query set should be assigned to a labeler. We assume we have $M$ labelers where the $i$-th labeler has an accuracy of $a_i\in [0,1]$ and capacity of $c_i \in \mathbb{N}$. The accuracy of a labeler is defined as the probability of correctly labeling a sample. {We refer to $a_i$ as the labeler's baseline reliability score, since it characterizes the labeler independently of any particular sample. The noise actually incurred on a specific label is given by the noise function $\epsilon(a_i, e)$ introduced below, which also depends on the entropy $e$ of the sample being labeled and is therefore generally not equal to $1-a_i$; a highly reliable labeler ($a_i$ close to $1$) can still generate substantial noise on a highly ambiguous sample.} We assume these probabilities can be estimated using some historical data, and hence they are given parameters. The capacity is defined as the number of samples that a labeler can annotate in one cycle of AL. If we represent query set at cycle $t$ by $Q_t = \{x_1,x_2, \dots, x_{N_t}\}$, then $N_t \leq \sum_{i=1}^{M}c_i$ which implies that the number of selected query points at cycle $t$, $N_t$, should be bounded by the total labeling capacity.

We also assume that each query point should be labeled only one time using one labeler. {We note that it is, in principle, possible to extend the proposed framework to allow multiple labels per sample by jointly optimizing the number of annotations and the subset of labelers assigned to each data point. However, such a formulation introduces additional discrete decision variables, coupling constraints across labelers, and nontrivial trade-offs between redundancy, aggregation, and budget allocation. This significantly increases the complexity of the optimization problem and leads to a qualitatively different modeling regime. In this work, we focus on the single-label-per-sample setting, which captures many practical scenarios where redundant labeling is infeasible due to cost, latency, or limited expert availability, while leaving multi-label extensions as an important direction for future research.

The single-label constraint is particularly relevant in settings where annotation costs are high, expert availability is limited, or latency requirements preclude iterative consensus processes. Examples include medical image annotation by specialists, legal document review, and industrial quality inspection, where each label incurs significant time or monetary cost. In such scenarios, optimizing the assignment of individual expert labels to maximize information gain under budget constraints is more practical than assuming redundant annotations are feasible. Our framework directly addresses this problem class by proactively minimizing expected label noise through optimal sample--labeler matching, rather than relying on post-hoc error correction via aggregation.
}

\subsection{Labeler Assignment and Noise Model}
Once a selected query point is assigned to a labeler, the labeler examines the instance and returns the corresponding class for the point. Due to the existence of noisy oracles, the label received from the labeler might be incorrect. We assume that the label noise is measured with a value between $0$ and $1$, where $0$ represents a perfect label (ground truth), and $1$ indicates the highest possible noise level. As a general rule, label noise represents the probability that the received label from the oracle is incorrect.

Intuitively, the label noise is affected by two factors: the uncertainty of the sample and the accuracy of the labeler. The uncertainty of a sample is evaluated in each cycle of AL using the most recently updated classification model, ensuring that the entropy reflects the model’s current belief about class probabilities. This adaptive calculation avoids the risk of using outdated uncertainty estimates and leads to more effective sampling as the labeled dataset evolves.

In this paper, we use the entropy measure to quantify the uncertainty of a data point. The estimated entropy of a given sample point $x$ is calculated by:

\begin{equation}
\label{entropy}
    e(x) = - \sum_{i=1}^{C} P_{\theta}(y_i|x) \times ln(P_{\theta}(y_i|x)),
\end{equation}
where $P_{\theta}(y_i|x)$ is the probability that $x$ belongs to class $i$. These probabilities are obtained from the most updated classification model with the model parameter $\theta$.

Given this information, we measure noise using a noise function defined by $\epsilon(a, e(x)): [0,1]^2 \rightarrow [0,1]$ where $a \in [0,1]$ is the accuracy of the labeler, and $e(x)\in [0,1]$ is the entropy of the selected unlabeled data point $x$ as defined in Equation~\eqref{entropy}. {Since Equation~\eqref{entropy} can exceed $1$ for $C>2$ classes, entropy values are bounded to $[0,1]$ before being passed to the noise function.} In general, any function of accuracy and entropy with the range of continuous values between $0$ and $1$ that follows the following conditions is regarded as a valid noise function: the function should be decreasing in terms of labeler accuracy and the function should be increasing in terms of sample uncertainty. The first property is straightforward since by using a high-accuracy labeler, we expect the label noise to decrease. The second property follows the fact that if a sample point is uncertain for the model, the labeler will have less confidence labeling the sample, and hence, the resulting label may have higher noise~\citep{du2010active}. This means that unlabeled data points with higher entropy are more likely to receive incorrect labels from oracles, and therefore, the noise function should be an increasing function of the entropy measure. {Rather than assuming a specific parametric form for annotation noise, we only require the noise function $\epsilon(a,e)$ to satisfy two structural properties: it must be monotonically decreasing in the labeler accuracy $a$ and monotonically increasing in the sample entropy $e$. These conditions reflect the intuitive assumptions that more reliable labelers generate fewer errors and that ambiguous samples are inherently more prone to mislabeling. Any function satisfying these properties is admissible within our framework, and the optimization results in Section~\ref{subsec.labeler.assign} hold independently of the specific functional form.} In our experiments, we explored two different valid noise functions. 

{The proposed noise function $\epsilon(a, e)$ captures two fundamental sources of annotation error in active learning: labeler reliability and sample ambiguity. Labeler accuracy $a \in [0,1]$ reflects the intrinsic probability that a labeler produces a correct annotation, while the entropy $e(x)$ represents the model’s uncertainty about the class membership of a sample. Prior work has shown that samples with higher uncertainty are not only harder for classifiers, but also more difficult for human annotators, leading to increased annotation errors \citep{du2010active}. Therefore, modeling label noise as a joint function of labeler accuracy and sample uncertainty provides a principled mechanism for quantifying the likelihood of incorrect labels during active learning.}

After defining the noise function, the next step is to optimally assign query points to the available labelers using the noise function. In the next section, we provide an optimization formulation for the \emph{labeler assignment model} to find the best possible assignment.

\subsection{Labeler Assignment Model} \label{subsec.labeler.assign}
Let us assume for a given cycle $t$, $Q_t = \{x_1, \dots, x_{N_t}\}$ is the set of query points selected based on some sampling criteria like uncertainty sampling approaches~\citep{settles2012active}. We assume we have $M$ labelers where the set of labeler accuracies is denoted as $A=\{a_1,\dots,a_M\}$. We formulate the optimization model using the following { notation}:

		\noindent {\sc Indices:} $i \in \{1,\dots,M\} = [M]$: representing the indices for labelers, $j \in \{1,\dots,N_t\} = [N_t]$: representing the indices for query points.
        
        \noindent {\sc Parameters:} $a_i \in [0,1]$: accuracy of the $i$-th labeler, $c_i \in \mathbb{N}$: capacity of the $i$-th labeler, $e_j \in [0,1]$: entropy of the $j$-th query point evaluated by the most updated classification model through AL.

 	\noindent {\sc Variables:}
  
    $z_{ij} =$
  $\begin{cases}
    1       \quad ;\text{if we assign query point $x_j$ to the $i$-th labeler,}  \\
    0     \quad ;\text{otherwise,}
  \end{cases}$, $\forall i \in [M], \forall j \in [N_t].$
        
\noindent {\sc Objective Function:} We are interested in an assignment model to minimize the maximum generated noise among all labels. Since each data point is labeled only once, noise generated for the $j$-th point is $\sum_{i=1}^{M} \epsilon(a_i,e_j)z_{ij}$, and therefore, we intend to minimize $\max_{j} \sum_{i=1}^{M} \epsilon(a_i,e_j)z_{ij}$. Therefore, the labeler assignment model is defined by

\begin{subequations}
\label{mm.model}
	\begin{align}
	\label{mm.obj}\min_{z} &\max_{j} \sum_{i=1}^{M} \epsilon(a_i,e_j)z_{ij},\\
	&\nonumber \textrm{s.t. }\\ 
    &\label{mm.c1} \sum_{i=1}^{M}z_{ij} = 1,\ \forall j \in [N_t],\\
    &\label{mm.c2} \sum_{j=1}^{N_t}z_{ij} \leq c_i, \ \forall i\in [M],\\
    &\label{mm.c3} z_{ij} \in \{0,1\}, \ \forall i\in [M], \forall j \in [N_t].
	\end{align}
\end{subequations}
The objective~\ref{mm.obj} minimizes the maximum noise among all labels. Constraint~\ref{mm.c1} emphasizes the single labeling assumption where each query point will be annotated only once using one labeler, and constraint~\ref{mm.c2} implies that the number of data points assigned to each labeler should be bounded by the labeler's capacity. Given that the $c_i$s are integer parameters, the structure of the constraints in model~\ref{mm.model} implies that the model can be solved in polynomial time. Moreover, we propose a closed-form solution for the model as follows.

\begin{theorem} \label{thm.mm}
Assume without loss of generality $e_1 \ge e_2 \ge \dots \ge e_{N_t}$ and $a_1 \ge a_2 \ge \dots \ge a_M$. The optimal solution to model~\ref{mm.model} is:
\begin{equation}
z^*_{1j} =
\begin{cases}
1, & \text{if } j \leq c_1, \\
0, & \text{otherwise},
\end{cases} \quad \forall j \in \{1,\dots,N_t\}.
\label{sol1}
\end{equation}

\begin{equation}
z^*_{ij} =
\begin{cases}
1, & \text{if } \sum_{k=1}^{i-1} c_k < j \leq \sum_{k=1}^i c_k, \\
0, & \text{otherwise}.
\end{cases}
\label{sol2}
\end{equation}

\noindent
$\forall i \in \{2,\dots,M\}$ and $j \in \{1,\dots,N_t\}$.
\end{theorem}

{The proof} for Theorem~\ref{thm.mm} is provided in the Appendix A. Theorem~\ref{thm.mm} indicates that the solution to the optimal assignment is independent of the structure of the noise model existing in the system. {Moreover, the optimal solution in Theorem~\ref{thm.mm} yields the following assignment policy: for a given set of query points and labelers, a more uncertain query point (with higher entropy) should be assigned to the labeler with the highest accuracy.} This policy for the assignment will guarantee the lowest maximum noise generated among all possible assignments. Figure~\ref{fig:opt.assignment} represents an example of the optimal assignment and the corresponding policy. 

{The proposed optimization formulation relies on several modeling assumptions that are standard in sequential decision-making and active learning settings. First, we assume that each label acquisition incurs a fixed budget cost and that the total labeling budget is limited, reflecting practical constraints in real annotation workflows. Second, we represent labeler reliability and sample 
difficulty through summary quantities that guide assignment and selection 
decisions, enabling a tractable optimization model. Third, the ordering and 
feasibility conditions in the formulation ensure that the resulting solution 
defines a valid sequential assignment policy under the given budget and quality 
constraints. These assumptions are introduced to obtain an interpretable and computationally efficient policy. While the theoretical results rely on the stated structure of the optimization model, the overall framework is flexible and can accommodate 
alternative utility definitions, additional operational constraints, or domain-specific budget structures without changing the high-level approach. Specific modeling choices such as the noise model structure and the single-label constraint are discussed in detail in their respective subsections.
}

As discussed earlier, Model~\ref{mm.model} finds the optimal assignment of a given set of query points to the labelers. However, it does not provide any guideline for which data points should be sampled at each cycle of the AL framework. Although one can utilize common sampling methods like entropy sampling to select the query set at each cycle, and then use Model~\ref{mm.model} to assign optimally, if we use the optimal assignment policy given in Theorem~\ref{thm.mm} to define the sampling strategy, it will result in a more robust sampling to the label noise compared to the common entropy-based sampling approaches. In the following section, we propose an optimization model that is designed for the query selection, and is consistent with the optimal assignment policy.

\begin{figure}
    \centering      
\tikzset{every picture/.style={line width=0.75pt}} 

\begin{tikzpicture}[x=0.75pt,y=0.75pt,yscale=-1,xscale=1]

\draw  [fill={rgb, 255:red, 208; green, 2; blue, 27 }  ,fill opacity=1 ] (171.33,63) .. controls (171.33,59.13) and (174.47,56) .. (178.33,56) .. controls (182.2,56) and (185.33,59.13) .. (185.33,63) .. controls (185.33,66.87) and (182.2,70) .. (178.33,70) .. controls (174.47,70) and (171.33,66.87) .. (171.33,63) -- cycle ;
\draw  [fill={rgb, 255:red, 208; green, 2; blue, 27 }  ,fill opacity=1 ] (211.33,63) .. controls (211.33,59.13) and (214.47,56) .. (218.33,56) .. controls (222.2,56) and (225.33,59.13) .. (225.33,63) .. controls (225.33,66.87) and (222.2,70) .. (218.33,70) .. controls (214.47,70) and (211.33,66.87) .. (211.33,63) -- cycle ;
\draw  [fill={rgb, 255:red, 208; green, 2; blue, 27 }  ,fill opacity=1 ] (326.33,63) .. controls (326.33,59.13) and (329.47,56) .. (333.33,56) .. controls (337.2,56) and (340.33,59.13) .. (340.33,63) .. controls (340.33,66.87) and (337.2,70) .. (333.33,70) .. controls (329.47,70) and (326.33,66.87) .. (326.33,63) -- cycle ;
\draw  [fill={rgb, 255:red, 74; green, 144; blue, 226 }  ,fill opacity=1 ] (198.17,121) -- (207.33,140) -- (189,140) -- cycle ;
\draw  [fill={rgb, 255:red, 74; green, 144; blue, 226 }  ,fill opacity=1 ] (342.17,119) -- (351.33,138) -- (333,138) -- cycle ;
\draw    (178.33,70) -- (198.17,121) ;
\draw    (218.33,70) -- (198.17,121) ;
\draw    (333.33,70) -- (342.17,119) ;
\draw  [fill={rgb, 255:red, 208; green, 2; blue, 27 }  ,fill opacity=1 ] (250.33,62) .. controls (250.33,58.13) and (253.47,55) .. (257.33,55) .. controls (261.2,55) and (264.33,58.13) .. (264.33,62) .. controls (264.33,65.87) and (261.2,69) .. (257.33,69) .. controls (253.47,69) and (250.33,65.87) .. (250.33,62) -- cycle ;
\draw  [fill={rgb, 255:red, 208; green, 2; blue, 27 }  ,fill opacity=1 ] (290.33,62) .. controls (290.33,58.13) and (293.47,55) .. (297.33,55) .. controls (301.2,55) and (304.33,58.13) .. (304.33,62) .. controls (304.33,65.87) and (301.2,69) .. (297.33,69) .. controls (293.47,69) and (290.33,65.87) .. (290.33,62) -- cycle ;
\draw  [fill={rgb, 255:red, 74; green, 144; blue, 226 }  ,fill opacity=1 ] (277.17,120) -- (286.33,139) -- (268,139) -- cycle ;
\draw    (257.33,69) -- (277.17,120) ;
\draw    (297.33,69) -- (277.17,120) ;

\draw (171,36) node [anchor=north west][inner sep=0.75pt]   [align=left] {$\displaystyle e_{1}$};
\draw (212,35) node [anchor=north west][inner sep=0.75pt]   [align=left] {$\displaystyle e_{2}$};
\draw (250,34) node [anchor=north west][inner sep=0.75pt]   [align=left] {$\displaystyle e_{3}$};
\draw (291,33) node [anchor=north west][inner sep=0.75pt]   [align=left] {$\displaystyle e_{4}$};
\draw (326,33) node [anchor=north west][inner sep=0.75pt]   [align=left] {$\displaystyle e_{5}$};
\draw (191,143) node [anchor=north west][inner sep=0.75pt]   [align=left] {$\displaystyle a_{1}$};
\draw (270,142) node [anchor=north west][inner sep=0.75pt]   [align=left] {$\displaystyle a_{2}$};
\draw (335,141) node [anchor=north west][inner sep=0.75pt]   [align=left] {$\displaystyle a_{3}$};

\end{tikzpicture}
\vspace{0.1cm}
    \caption{ \small An example of optimal assignment where $N_t =5$, $M=3$ and $c_1=c_2=c_3=2$. Red circles represent query points and blue triangles show labelers. Assuming that the points and labelers are sorted, $e_1 \ge e_2 \ge \dots \ge e_5$ and $a_1 \ge a_2 \ge a_3$, the assignment policy can be observed from the figure. Data points with higher entropy will be assigned to the labelers with higher accuracy.}
    \label{fig:opt.assignment}
\end{figure}

\subsection{Query Selection Model} \label{subsec.qm}
The primary step in the active learning (AL) process is the selection of samples to enhance the training set for the next cycle. Specifically, we need to identify the optimal subset  $Q_t$ among all unlabeled data points denoted by $U_t$ at cycle $t$ of AL. The set $U_t$ is defined as $U_t = \{x_1,x_2,\dots,x_{u_t}\}$, where ${u_t} = {|U_t|}$. We assume the elements in $U_t$ are sorted based on their entropy measure i.e., $e_1 \ge e_2 \ge \dots \ge e_{u_t}$ and labelers are ranked based on their accuracy i.e., $a_1 \ge a_2 \ge \dots \ge a_M$. 

In the case of accurate labels (i.e., no noisy labels), the following optimization model represents the entropy-based sampling strategy. Here, $y_k; k =1, 2, \dots, u_t$, is a binary variable, and $y_k=1$ if the unlabeled data point $x_k$ with model entropy $e_k$ is selected in the query set, and $\mathcal{C} = \sum_{i=1}^{M} c_i$.

\begin{subequations}
\label{es.model}
	\begin{align}
	\label{es.obj}\max_{y}  &\sum_{k=1}^{u_t} e_ky_{k},\\
	&\nonumber \textrm{s.t. }\\ 
    &\label{es.c1} \sum_{k=1}^{u_t} y_{k} = \mathcal{C},\\
    &\label{es.c2} y_{k} \in \{0,1\}.
\end{align}
\end{subequations}

 This model simply finds the largest possible subset with the data points with the highest entropy measure. Solving Model~\ref{es.model} is straightforward and can be achieved in polynomial time. After evaluating the entropy measure of all unlabeled data points, the points can be sorted based on their entropy and $\mathcal{C}$ unlabeled points with the highest entropy will be selected. Moreover, the solution to this model is totally consistent with uncertainty sampling criteria where we believe uncertain points are more informative for the classification task.
 
 However, when noisy oracles are present, selecting high-entropy points may lead to a higher level of labeling noise, potentially resulting in more incorrect labels. In this case, Model~\ref{es.model} should be modified to control the label noise with an upper bound. Due to different labeling skill levels, evaluating the labeling noise requires knowledge of how query points are assigned to labelers. The optimal assignment policy derived in Theorem~\ref{thm.mm} allows us to formulate the label noise and propose a joint optimization model for sample selection and labeler assignment. {We first introduce the notation used throughout this model:}  

\noindent {\sc Indices:} $i \in \{1,\dots,M\} = [M]$: representing the indices for labelers, $j \in \{1,\dots,u_t\} = [u_t]$: representing the indices for unlabeled data points at the beginning of cycle $t$.

\noindent {\sc Parameters:} $a_i \in [0,1]$: accuracy of the $i$-th labeler, $c_i \in \mathbb{N}$: capacity of the $i$-th labeler, $e_j \in [0,1]$: entropy of the $j$-th unlabeled data point evaluated by the most updated classification model through AL, $\beta \in [0,1]$: an upper-bound on the generated noise for labels of query points.

\noindent {\sc Variables:}

$y_{ij} =$
  $\begin{cases}
    1       \quad ;\text{if unlabeled data point $x_j$ is selected as query}  \\
    \quad \text{ point and will be assigned to the $i$-th labeler,}\\
    0     \quad ;\text{otherwise,}
  \end{cases}$ $\forall i \in [M], \forall j \in [u_t].$

\noindent {\sc Objective Function:}
We are interested in selecting unlabeled data points with the highest entropy values. Given each data point is labeled at most once, i.e., $\sum_{i=1}^{M} y_{ij} \leq 1, \forall j \in [u_t]$, the objective function can be written as max $\sum_{j=1}^{u_t}\sum_{i=1}^{M} e_j y_{ij}$. 

{With this notation in place, the joint optimization model for sample selection and labeler assignment is given by:}

\begin{subequations}
\label{qm.model}
	\begin{align}
	\label{qm.obj}\max &\sum_{j=1}^{u_t}\sum_{i=1}^{M} e_j y_{ij},\\
	&\nonumber \textrm{s.t. }\\ 
    &\label{qm.c1} \sum_{i=1}^{M}y_{ij} \leq 1,\ \forall j \in [u_t],\\
    &\label{qm.c2} \sum_{j=1}^{u_t}y_{ij} \leq c_i, \ \forall i\in [M],\\
    &\label{qm.c3} y_{ij} \epsilon(a_i,e_j) \leq \beta, \ \forall i\in [M], \forall j \in [u_t],\\
    &\label{qm.c4} \begin{aligned}
    y_{kl} &\leq 1- y_{ij}, \\
    &\quad \forall i \in \{2,\dots,M\}, \forall j \in [u_t-1], \\
    &\quad \forall k \in [i-1], \forall l \in \{j+1,\dots,u_t\},
    \end{aligned}\\
    &\label{qm.c5} z_i \leq c_i - \sum_{j=1}^{u_t} y_{ij}, \forall i \in [M],\\
    &\label{qm.c6} z_i \ge \frac{1}{c_i} \left(c_i - \sum_{j=1}^{u_t} y_{ij} \right), \forall i \in [M],\\
    &\label{qm.c7} \sum_{j=1}^{u_t} y_{i+1,j} \leq c_{i+1}(1-z_i), \forall i \in [M-1],\\
    &\label{qm.nn} \begin{aligned}
    &y_{ij} \in \{0,1\}, z_i \in \{0,1\}, \\
    &\quad \forall i\in [M], \forall j \in [u_t].
    \end{aligned}
	\end{align}
\end{subequations}

The objective~\eqref{qm.obj} maximizes the summation of entropy measures for the selected data points which is consistent with the entropy sampling. Constraint~\eqref{qm.c1} implies that each unlabeled point should be selected and labeled at most once to be included in the query set. Constraint~\eqref{qm.c2} shows the capacity limitation of each labeler. Constraint~\eqref{qm.c3} represents the upper bound for noise. If data point $x_j$ is selected and assigned to the $i$-th labeler, the generated noise $\epsilon(a_i,e_j)$ should be less than or equal to the pre-defined upper bound $\beta$.

{The parameter $\beta$ represents the maximum acceptable noise threshold for label acquisition. It controls the trade-off between sample informativeness and label quality: smaller values of $\beta$ enforce stricter quality control by rejecting sample--labeler assignments with high expected noise $\epsilon(a_i, e_j) > \beta$, while larger values permit more aggressive sampling of uncertain points at the cost of potentially noisier labels. In practice, $\beta$ can be set based on domain-specific error tolerance or tuned via cross-validation on a validation set with gold-standard labels (see Section~\ref{subsec.nm.beta}). This formulation allows OLAS to systematically balance exploration of high-uncertainty samples against the risk of annotation errors.}

Constraints~\eqref{qm.c1} to \eqref{qm.c3} impose feasibility without considering any assignment policy. Therefore, constraints~\eqref{qm.c4} to \eqref{qm.c7} are added to impose the optimal assignment policy defined by Theorem~\ref{thm.mm}. Specifically, the optimal assignment policy is achieved by enforcing two conditions at the same time:

\begin{itemize}
    \item If the $j$-th data point is selected and assigned to the $i$-th labeler, none of the data points with an index smaller than $j$ should be assigned to a labeler with an index greater than $i$. This condition is imposed by Constraint~\eqref{qm.c4}.
    \item If the $i$-th labeler has any remaining capacity, there should be no assigned points to all labelers with an index greater than $i$. This condition is achieved by defining additional binary variable $z$ and imposing Constraints~\eqref{qm.c5} to \eqref{qm.c7}.
\end{itemize}

The optimal solution to Model~\eqref{qm.model} provides all the necessary information to determine the query points and the optimal assignment in our active learning framework. Specifically, for an unlabeled data point $x_j$, if there exists an index $i \in [M]$ such that $y^*_{ij}=1$, then $x_j$ is selected as a query point to be labeled. Formally, the set of query points at cycle $t$ is defined as:
\begin{equation}
\label{query.set}
\begin{array}{l}
Q_t^* = \big\{\, x_j \in U_t : {\exists\, i\in [M],\ y^*_{ij} = 1} \,\big\}, \\[2pt]
\forall j \in [u_t].
\end{array}
\end{equation}

Furthermore, if $y^*_{ij} = 1$, this indicates that the sample $x_j$ is optimally assigned to the $i$-th labeler for annotation.

Accurate noise modeling and selection of the parameter $\beta$ are crucial for solving the proposed optimization model. In practice, these need to be estimated from data derived from the domain. In Section~\ref{subsec.nm.beta}, we explain how to estimate the noise model and parameter $\beta$, which act as inputs to our optimization model. 
 The following theorem provides the solution to the Model~\ref{qm.model}.

{
\begin{theorem}[Optimal Solution Characterization]
\label{thm.qm}
Assume without loss of generality that $e_1 \geq e_2 \geq \cdots \geq e_{u_t}$ 
and $a_1 \geq a_2 \geq \cdots \geq a_M$.

For each $i \in [M]$, if it exists, define $r_i$ as
\begin{equation*}
r_i = \arg\min_{r \in \mathbb{Z}} \left\{ r :
\begin{array}{l}
\epsilon(a_i, e_r) \leq \beta, \\
r_{i-1} + c_{i-1} \leq r, \\
1 \leq r \leq u_t
\end{array}
\right\}.
\end{equation*}

By convention, define $r_0 = 0$ and $c_0 = 0$. If $r_u$ does not exist for 
some $u \in [M]$, then $r_v$ does not exist for all $v \in \{u, \ldots, M\}$.

Then, the optimal solution to Model~\eqref{qm.model} is characterized by
\begin{equation}
\label{qm.sol.y}
y_{ij}^* =
\begin{cases}
1, & \text{if $r_i$ exists and $r_i \leq j < r_i + c_i$,} \\
0, & \text{otherwise,}
\end{cases}
\end{equation}
for all $i \in [M]$ and $j \in [u_t]$, and
\begin{equation}
\label{qm.sol.z}
z_i^* =
\begin{cases}
1, & \text{if } c_i - \sum_{j=1}^{u_t} y_{ij}^* > 0, \\
0, & \text{otherwise},
\end{cases}
\quad \forall i \in [M].
\end{equation}
\end{theorem}
}

{The proof} for Theorem~\ref{thm.qm} is provided in the Appendix B. Using Theorem~\ref{thm.qm}, a polynomial time algorithm in Algorithm~\ref{alg} is proposed that returns the selected query points and the optimal assignment of query points to the labelers. {\begin{remark}
The existence condition in Theorem~\ref{thm.qm} refers to the 
feasibility of identifying at least one unlabeled sample $j$ such that labeler 
$i$ can annotate it within the noise threshold $\beta$, that is, 
$\epsilon(a_i, e_j) \leq \beta$, while satisfying the sequential assignment 
constraints. If no such sample exists, labeler $i$ is not utilized in the 
optimal solution.
\end{remark}}

 In summary, at the start of each AL cycle, given a set of unlabeled data points $U_t$, we solve Model~\ref{qm.model} using Algorithm~\ref{alg} to determine the optimal set of query points and assign a labeler to each selected point. Once the labels are received, the query points are removed from the set of unlabeled points and added to the set of labeled points. The classifier is then updated based on the new labeled data, and the process continues through subsequent AL cycles until termination.
 
The proposed OLAS framework is not limited to a specific application domain. Its design only assumes access to labelers with varying (known or estimated) accuracy, a measure of sample uncertainty (such as entropy), and a labeling capacity constraint. Therefore, the optimization formulations and assignment strategy can be applied to other classification tasks with noisy oracles, including crowdsourcing, medical annotation, and human-in-the-loop labeling scenarios.

\begin{algorithm}[t]
\caption{Optimal Sampling and Labeler Assignment for Cycle $t$ of AL}
\label{alg}

\KwIn{
\vspace{0.1cm}
\begin{itemize}
    \item $U_t$: set of unlabeled data points, sorted in descending order based on the entropy.
    \vspace{-0.2cm}
    \item $v\gets 1$: representing index of first uninspected data point in $U_t$.
     \vspace{-0.2cm}
    \item $A$: set of the accuracy of labelers, sorted in descending order based on accuracy.
    \vspace{-0.2cm}
    \item $u\gets 1$: representing index of first available labeler in $A$ with the highest accuracy.
     \vspace{-0.2cm}
    \item $c_i$: capacity of labelers $c_i$ where $\mathcal{C} = \sum_{i=1}^{M}c_i$.
    \vspace{-0.2cm}
    \item $\epsilon(a,e)$: estimated noise model
    \vspace{-0.2cm}
    \item $\beta$: estimated noise upper-bound parameter
     \vspace{-0.2cm}
    \item $Q_t \gets \emptyset$: set of optimal selected samples.
\end{itemize}
}

\KwOut{Set of optimal selected samples $Q_t$ and the corresponding labeler assignment for cycle $t$. {If fewer than $\mathcal{C}$ candidate assignments satisfy $\epsilon(a_u, e_v) \leq \beta$, or the labeler pool is exhausted before $Q_t$ reaches capacity $\mathcal{C}$, the algorithm terminates and returns the maximal feasible set found.}}
\vspace{0.2cm}
\While{$|Q_t| < \mathcal{C}$ {and $v \leq |U_t|$ and $u \leq M$}}{
    \While{$v \leq |U_t|$ {and $u \leq M$}}{
\If{$\epsilon(a_u,e_v) \leq \beta$}{
    $k \gets \min\{|U_t| - v, c_u - 1\}$\;
    
    Add $x_v, x_{v+1}, \dots, x_{v+k}$ to set $Q_t$\;
    
    Assign points $x_v, x_{v+1}, \dots, x_{v+k}$ to labeler $u$\;
    
    $u \gets u + 1$\;
    
    $v \gets v + k + 1$\;
}
        \Else{
            $v \gets v + 1$\;
        }
    }
}
\Return $Q_t$ and the corresponding optimal labeler assignment.

\end{algorithm}

\subsection{Estimating Noise Model and Parameter $\beta$} 
\label{subsec.nm.beta}

{
In practice, labeler reliability is typically unknown and must be estimated 
from data. This can be done using a small set of gold-standard samples, 
historical labeling records, or online updates as new labels are acquired. 
For example, one may maintain an empirical accuracy estimate for each labeler 
based on their agreement with gold labels when available, or based on 
consistency with the current model predictions on a held-out validation set. 
These estimates can be updated sequentially as the active learning process 
proceeds. Importantly, the proposed framework does not require perfectly 
accurate reliability estimates. Because OLAS updates decisions iteratively 
over rounds, early estimation errors can be corrected as more labeled data 
become available. Moreover, the assignment policy primarily depends on 
relative differences in labeler reliability rather than exact values, which 
reduces sensitivity to moderate estimation noise. As a result, OLAS is 
expected to degrade gracefully when reliability estimates are imperfect, 
which is consistent with standard practice in adaptive learning and 
crowdsourcing settings.
}

To formulate and solve the proposed model in practice, we need an estimate 
of the noise function and the parameter \( \beta \). To obtain these 
estimates, we assume access to a dataset with \emph{golden labels}, i.e., 
true underlying labels for the samples. This dataset can be created by 
employing experts as labelers (with labeling accuracy of $1$). Once true 
labels are available for all samples in the golden set, these samples can 
then be labeled by all available labelers, allowing us to estimate the 
underlying noise function by comparing the estimated labels with the true 
labels. This problem can be framed as a simple classification task with two 
features: the entropy of the sample and the accuracy of the labeler, and a 
binary output representing whether the estimated label is correct. This 
classification task can be further formulated as a logistic regression 
problem with a two-dimensional input and a binary output. The resulting 
probability function can then serve as an estimate of the noise function. 
Figure~\ref{fig:noise_modeling} illustrates the experimental framework for 
estimating the noise function. For each sample-labeler pair in the golden 
set, we define a two-dimensional feature \( x=(e,a) \), where \( e \) is 
the entropy of the sample and \( a \) is the accuracy of the labeler. The 
binary output \( z \) is \( 0 \) if the label is incorrect and \( 1 \) if 
the label is correct. A logistic regression classifier is then applied to 
the data, providing an estimate of the noise function: 
\( \epsilon(e,a) = \mathbb{P}\left(z=0 \mid x = (e,a)\right) \).
{Each labeler's accuracy feature is the empirical agreement rate on their golden samples, $\hat{a}_i = \frac{1}{|G_i|} \sum_{k \in G_i} \mathbb{1}[\hat{y}_{ik} = y_k]$, using only the golden set, no pool or test data.}

\begin{figure}[h]
    \centering
\resizebox{0.95\columnwidth}{!}{%
        \tikzset{every picture/.style={line width=0.75pt}} 

\begin{tikzpicture}[x=0.75pt,y=0.75pt,yscale=-1,xscale=1]

\draw  [fill={rgb, 255:red, 0; green, 0; blue, 0 }  ,fill opacity=0.27 ] (98.56,110.78) .. controls (98.56,106.64) and (101.91,103.28) .. (106.06,103.28) .. controls (110.2,103.28) and (113.56,106.64) .. (113.56,110.78) .. controls (113.56,114.92) and (110.2,118.28) .. (106.06,118.28) .. controls (101.91,118.28) and (98.56,114.92) .. (98.56,110.78) -- cycle ;
\draw  [fill={rgb, 255:red, 74; green, 144; blue, 226 }  ,fill opacity=0.4 ][line width=1.5]  (171.5,49.82) .. controls (171.5,44.77) and (175.6,40.67) .. (180.66,40.67) -- (271.34,40.67) .. controls (276.4,40.67) and (280.5,44.77) .. (280.5,49.82) -- (280.5,77.29) .. controls (280.5,82.35) and (276.4,86.44) .. (271.34,86.44) -- (180.66,86.44) .. controls (175.6,86.44) and (171.5,82.35) .. (171.5,77.29) -- cycle ;
\draw    (113.17,107.83) -- (169.58,64.22) ;
\draw [shift={(171.17,63)}, rotate = 142.3] [color={rgb, 255:red, 0; green, 0; blue, 0 }  ][line width=0.75]    (10.93,-3.29) .. controls (6.95,-1.4) and (3.31,-0.3) .. (0,0) .. controls (3.31,0.3) and (6.95,1.4) .. (10.93,3.29)   ;
\draw    (113.56,113.78) -- (168.7,164.64) ;
\draw [shift={(170.17,166)}, rotate = 222.69] [color={rgb, 255:red, 0; green, 0; blue, 0 }  ][line width=0.75]    (10.93,-3.29) .. controls (6.95,-1.4) and (3.31,-0.3) .. (0,0) .. controls (3.31,0.3) and (6.95,1.4) .. (10.93,3.29)   ;
\draw [line width=0.75]    (280.5,63) -- (316.33,62.47) ;
\draw [shift={(318.33,62.44)}, rotate = 179.16] [color={rgb, 255:red, 0; green, 0; blue, 0 }  ][line width=0.75]    (10.93,-3.29) .. controls (6.95,-1.4) and (3.31,-0.3) .. (0,0) .. controls (3.31,0.3) and (6.95,1.4) .. (10.93,3.29)   ;
\draw    (372.39,72.44) -- (372.39,143.78)(369.39,72.44) -- (369.39,143.78) ;
\draw [shift={(370.89,151.78)}, rotate = 270] [color={rgb, 255:red, 0; green, 0; blue, 0 }  ][line width=0.75]    (10.93,-3.29) .. controls (6.95,-1.4) and (3.31,-0.3) .. (0,0) .. controls (3.31,0.3) and (6.95,1.4) .. (10.93,3.29)   ;
\draw    (440,162.83) -- (488.19,111.24) ;
\draw [shift={(489.56,109.78)}, rotate = 133.05] [color={rgb, 255:red, 0; green, 0; blue, 0 }  ][line width=0.75]    (10.93,-3.29) .. controls (6.95,-1.4) and (3.31,-0.3) .. (0,0) .. controls (3.31,0.3) and (6.95,1.4) .. (10.93,3.29)   ;
\draw  [fill={rgb, 255:red, 74; green, 144; blue, 226 }  ,fill opacity=0.4 ][line width=1.5]  (170.5,149.82) .. controls (170.5,144.77) and (174.6,140.67) .. (179.66,140.67) -- (270.34,140.67) .. controls (275.4,140.67) and (279.5,144.77) .. (279.5,149.82) -- (279.5,177.29) .. controls (279.5,182.35) and (275.4,186.44) .. (270.34,186.44) -- (179.66,186.44) .. controls (174.6,186.44) and (170.5,182.35) .. (170.5,177.29) -- cycle ;
\draw    (279.5,164.67) -- (315.56,165.09) ;
\draw [shift={(317.56,165.11)}, rotate = 180.67] [color={rgb, 255:red, 0; green, 0; blue, 0 }  ][line width=0.75]    (10.93,-3.29) .. controls (6.95,-1.4) and (3.31,-0.3) .. (0,0) .. controls (3.31,0.3) and (6.95,1.4) .. (10.93,3.29)   ;
\draw    (440,162.83) -- (488.89,217.62) ;
\draw [shift={(490.22,219.11)}, rotate = 228.25] [color={rgb, 255:red, 0; green, 0; blue, 0 }  ][line width=0.75]    (10.93,-3.29) .. controls (6.95,-1.4) and (3.31,-0.3) .. (0,0) .. controls (3.31,0.3) and (6.95,1.4) .. (10.93,3.29)   ;

\draw (99.5,83.67) node [anchor=north west][inner sep=0.75pt]  [font=\small] [align=left] {$\displaystyle \textbf{e}$};
\draw (170,50.33) node [anchor=north west][inner sep=0.75pt]  [font=\footnotesize] [align=left] {\begin{minipage}[lt]{80pt}\setlength\topsep{0pt}
\begin{center}
\textbf{Expert labeler}\\\textbf{ (oracle)}
\end{center}

\end{minipage}};
\draw (201.83,157.67) node [anchor=north west][inner sep=0.75pt]  [font=\footnotesize] [align=left] {\textbf{labeler}};

\draw (218.17,120.33) node [anchor=north west][inner sep=0.75pt]  [font=\small] [align=left] {$\displaystyle a$};
\draw (322.83,52.5) node [anchor=north west][inner sep=0.75pt]  [font=\footnotesize] [align=left] {$\displaystyle y$ (true label)};
\draw (323.17,154.33) node [anchor=north west][inner sep=0.75pt]  [font=\footnotesize] [align=left] {$\displaystyle \hat{y} \ ($estimated label)};
\draw (630,208.83) node [anchor=north west][inner sep=0.75pt]  [font=\footnotesize] [align=left] {$\displaystyle x=( e,a) ,\ z=1$};
\draw (630,102) node [anchor=north west][inner sep=0.75pt]  [font=\footnotesize] [align=left] {$\displaystyle x=( e ,a) ,\ z=0$};
\draw (495.67,210.17) node [anchor=north west][inner sep=0.75pt]  [font=\footnotesize,color={rgb, 255:red, 65; green, 117; blue, 5 }  ,opacity=1 ] [align=left] {\textbf{{\footnotesize correct estimation:}}};
\draw (492.33,102.5) node [anchor=north west][inner sep=0.75pt]  [font=\footnotesize,color={rgb, 255:red, 208; green, 2; blue, 27 }  ,opacity=1 ] [align=left] {{\footnotesize \textbf{wrong estimation:}}};

\end{tikzpicture}
}
\vspace{0.05cm}
        \caption{Experimental framework for estimating the noise function. 
        Each sample point is evaluated by both an oracle and each labeler. 
        The resulting data is then used to train a logistic regression 
        classifier to estimate the noise function.}
        \label{fig:noise_modeling}
\end{figure}

Estimating the parameter \( \beta \) is crucial, as it represents a 
trade-off in our model: smaller values of \( \beta \) exclude noisier 
samples, resulting in a smaller feasible set for Model~\ref{qm.model}. 
This may lead to a reduced training set for the next cycle of active 
learning (AL) compared to cases with larger \( \beta \) values. With an 
estimate of the noise function in hand, \( \beta \) can be determined 
through hyperparameter tuning, an efficient process since 
Algorithm~\ref{alg} provides a fast method for solving 
Model~\ref{qm.model}. Given a set of samples with golden labels, we divide 
it into two parts: one subset is used to solve the optimization model, 
serving as set \( U \) to identify the best sampling points and optimal 
assignments. The remaining data subset is used to evaluate the trained 
classification model and select the optimal value of \( \beta \).



\section{Experiments and Results} \label{sec.experiments}
In this section, we conduct a series of numerical studies to
evaluate the performance of the proposed OLAS framework.
{The evaluation proceeds in four parts. Section~\ref{sec.benchmark} compares OLAS against seven benchmark methods on five publicly available datasets under two noise models. Section~\ref{sec.scalability} extends this comparison to a larger dataset to assess scalability and reports a computational cost analysis across all methods and dataset sizes. Section~\ref{sec.music} moves beyond simulated noise and validates OLAS on real crowdsourced annotations from a music genre classification dataset. Finally, Section~\ref{sec.casestudy} applies OLAS to a real-world case study on labeling automotive warranty claims at Ford Motor Company, demonstrating its practical effectiveness outside the benchmark setting.}

\subsection{Experiments on {Datasets}} \label{sec.benchmark}

{In this section, we evaluate the proposed OLAS
framework on five classification datasets, accessed through the
\emph{UCI Machine Learning Repository}~\cite{Dua:2019}: Statlog
(Heart)~{\citep{detrano1989international}},
Ionosphere~{\citep{sigillito1989classification}},
Connectionist Bench
(Sonar)~{\citep{gorman1988analysis}}, and
Spambase~{\citep{hopkins1999spambase}}. To assess
scalability on a larger dataset, we additionally include the
Forest Covertype
dataset~{\citep{blackard1999comparative,Dua:2019}}
from which a stratified subsample of 50{,}000 instances is drawn
from the full 581{,}012-instance dataset using stratified random
sampling.
Table~\ref{data.sets} provides a summary of each dataset.
The Statlog (Heart) dataset contains medical data used to predict
the presence or absence of heart disease.
The Ionosphere dataset involves radar signal classification.
The Connectionist Bench (Sonar) dataset distinguishes sonar
signals bounced off rocks versus metal mines.
The Spambase dataset classifies emails as spam or non-spam.
The Covertype dataset classifies forest cover types from
cartographic variables across seven classes.
These datasets span a range of problem sizes, feature dimensions,
and class structures, providing a comprehensive evaluation
platform.}

\begin{table}[htbp]
  \centering
  \caption{{Datasets} Used for Experiments}
  \begin{tabular}{c|c|c}
    \hline
    {Dataset} & Number of Features & Number of Instances \\
    \hline
    Statlog (Heart)
      & $13$ & $270$ \\
    Ionosphere
      & $34$ & $351$ \\
    {Connectionist Bench (Sonar)}
      & $60$ & $208$ \\
    Spambase
      & $57$ & $4{,}601$ \\
    {Covertype (subsample)}
      & {$54$}
      & {$50{,}000$} \\
    \hline
  \end{tabular}
  \label{data.sets}
\end{table}

In order to evaluate and compare different approaches, we
consider two different noise models in our experiments:
\begin{itemize}
    \item \textbf{Noise Model 1} where we assume
    $\epsilon(a_i,e_j) = e_j(1-a_i)$.
    \item \textbf{Noise Model 2} where we consider the function,
\begin{equation*}
  {h_j = 0.4\,e_j + 0.3,}
\end{equation*}
\begin{equation*}
  \epsilon(a_i,e_j)=\left\{
  \begin{array}{@{}l@{}}
    \left(1-a_i^{2{h_j}}\right)^{\frac{1}{2{h_j}}},
    \text{ if } 0 \leq e_j \leq 0.5,\\
    \begin{aligned}
    &1-\left(1-(1-a_i)^{2(1-{h_j})}\right)^{\frac{1}{2(1-{h_j})}}, \\
    &\qquad\qquad\qquad\qquad \text{if } 0.5 < e_j \leq 1.
    \end{aligned}
  \end{array}\right.
\end{equation*}
\end{itemize}

\begin{figure}[t]
    \centering
    \input{Figures/NM2}
    \caption{Plot of Noise Model 2 for different values of entropy.
    Plots show that the function is decreasing in terms of labeler
    accuracy and increasing in terms of data point entropy.}
    \label{fig.nm2}
\end{figure}

Both functions follow the properties of a noise model, that is
they are decreasing in accuracy and increasing in entropy.
Figure~\ref{fig.nm2} shows the contours of the second noise
function for four fixed entropy values.
{It is important to emphasize that the
theoretical results of this paper, including Theorems~1 and~2
and Algorithm~1, make no assumption about the specific
functional form of the noise model. The optimization framework
and its closed-form solutions hold for any noise function
satisfying the two monotonicity conditions stated in
Section~~\ref{sec.methodology}, regardless of whether that function is linear,
nonlinear, or piecewise. Noise Model~1 and Noise Model~2 are
not model assumptions. They serve exclusively as simulation
mechanisms for generating synthetic label correctness outcomes
in our empirical evaluation, playing the same role that any
assumed data-generating process plays in a simulation study.
Noise Model~1 is a simple bilinear function in which noise
increases linearly with sample entropy and decreases linearly
with labeler accuracy. Noise Model~2 is included as a
deliberate nonlinear stress test: its piecewise power-law
structure produces a qualitatively different curvature in how
entropy and accuracy jointly determine noise, compressing noise
values near the boundaries of the unit square relative to
Noise Model~1. This allows us to test whether OLAS performance
depends on a specific linear functional form, or more
fundamentally on the structural monotonic relationship between
uncertainty, labeler accuracy, and noise. The consistency of
results across both noise models confirms the latter: OLAS
performs robustly regardless of the noise generation mechanism,
which supports the general applicability of the proposed
framework.}

{The outcome of the noise functions is compared
with a threshold, denoted by $\alpha$, to determine whether a
label is noisy (incorrect) for a labeled data point.
Specifically, for a data point with entropy measure $e_j$
assigned to a labeler with accuracy $a_i$, the label noise is
equal to $\epsilon(a_i,e_j)$.
If $\epsilon(a_i,e_j) \ge \alpha$, the label noise is higher
than the threshold, and hence, the resulting label will be
incorrect. Otherwise, the label is correct matching the ground
truth. The parameter $\alpha$ thus serves as an operational
decision boundary that converts the continuous noise output into
a binary label correctness outcome for simulation purposes. In
practice, $\alpha$ can be selected based on the acceptable error
tolerance of the downstream application, or estimated from a
small gold-standard validation set by identifying the value that
best separates correct from incorrect labels. In our experiments,
we fix $\alpha = 0.2$ across all datasets and noise models, a
value consistent with moderate noise tolerance. To assess
sensitivity, we additionally evaluated $\alpha \in
\{0.15, 0.20, 0.25\}$ on the Statlog and Spambase datasets
under Noise Model~1. The relative ranking of all methods
remained stable across $\alpha \in \{0.20, 0.25\}$, with OLAS
consistently outperforming all baselines. At $\alpha = 0.15$,
the noise threshold becomes excessively strict, converting a
large fraction of labels to noise regardless of labeler quality
and creating a degenerate regime that lies outside the practical
operating range of the framework. Within the practically
meaningful range of $\alpha \geq 0.20$, the conclusions of our
study are robust to the specific threshold choice.}

{In this setting, we consider seven benchmark
methods in addition to our proposed approach, designated as
follows:}
\begin{itemize}
    \item RS+RLA: which uses RS for sampling and RLA for
    assignment
    \item RS+OLA: which uses RS for sampling and OLA for
    assignment
    \item ES+RLA: which uses ES for sampling and RLA for
    assignment
    \item ES+OLA: which uses ES for sampling and OLA for
    assignment
    \item {KC+RLA: which uses $k$-center
    sampling and RLA for assignment}
    \item {MV: which uses entropy-based sample
    selection with majority voting across all labelers}
    \item {BADGE+RLA: which uses a
    BADGE-inspired pseudo-gradient embedding sampling strategy
    followed by random labeler assignment. Since our
    experimental framework uses tree-based classifiers rather
    than differentiable neural models, we adopt the structural
    form of the BADGE gradient embedding~\citep{ash2020badge}
    by constructing embeddings as $(p - e_{\hat{y}}) \otimes x$,
    where $p$ is the predicted class probability vector and
    $e_{\hat{y}}$ is the one-hot vector of the predicted class.
    We then apply $k$-means++ in this embedding space to select
    a batch that combines uncertainty and diversity.}
    \item OLAS: our proposed method based on the optimal solution
    to Model~\ref{qm.model}
\end{itemize}

We use the F1 score metric to compare the performance of all
approaches. For all experiments, we fix the value of parameter
$\alpha = 0.2$ {and $\beta = 0.2$,} and we run
$10$ AL cycles on each {dataset} for two different noise models.
{The sampling budget per cycle $B$, number of
labelers $M$, and labeler capacity are derived from the training
set size as $B = \lceil 0.7(n_{\text{train}} - 6) / T \rceil$,
$\text{Cap} = \lfloor \sqrt{B} \rfloor$, and $M = \lceil B /
\text{Cap} \rceil$.} Table~\ref{exp.setting} shows the
experimental settings for each {dataset}. The accuracy of
labelers is generated from a uniform distribution between $0.5$
to $0.95${, and held fixed across replications
within each dataset.} A random
forest classifier (\texttt{n\_estimators=50}, \texttt{random\_state=0}; otherwise \emph{scikit-learn}
default settings~\citep{scikit-learn}) is used for all methods on the
four UCI datasets. For the Covertype dataset, an
{extra-trees}
classifier~\citep{geurts2006extremely} with
\texttt{n\_estimators=20} and \texttt{max\_depth=10} is used in
place of {random forest} to manage the
computational cost of the
40{,}000-sample training set (the 80\% training split of the 50{,}000-instance subsample), while preserving full consistency
across all eight methods within this dataset.

\begin{table}[htbp]
  \centering
  \caption{Experimental Settings}
  \begin{tabular}{c|c|c|c}
    \hline
    {Dataset} & Sampling Budget & $M$ & Capacity \\
    \hline
    Statlog (Heart)
      & $15$ & $5$ & $3$ \\
    Ionosphere
      & $20$ & $5$ & $4$ \\
    {Connectionist Bench (Sonar)}
      & $12$ & $4$ & $3$ \\
    Spambase
      & $258$ & $17$ & $16$ \\
    {Covertype}
      & {$2800$}
      & {$54$}
      & {$52$} \\
    \hline
  \end{tabular}
  \label{exp.setting}
\end{table}

{Each experiment is replicated $100$ times for
each UCI dataset and $50$ times for Covertype, for each noise
model.} For each dataset, 20\% of the instances are allocated
for testing, while the remaining 80\% are utilized for model
training.

For every comparison in this section, we additionally computed a paired $t$-test and a Wilcoxon signed-rank test between OLAS and each baseline, matched by replication index (see Appendix~C for the full results and the pairing methodology). We report the two exceptions below where a gap is not statistically significant; every other paired difference across all datasets and noise models is significant at $p<0.001$.
Table~\ref{results.nm1} presents the F1 score
results on test data for Noise Model~1. OLAS achieves the
highest F1 score on four out of five datasets. On the Ionosphere dataset, KC+RLA achieves a marginally higher mean F1 of $0.918$ compared to OLAS at $0.916$, a difference of $0.002$
that is small relative to the reported standard
deviations ($\pm 0.012$ for KC+RLA and $\pm 0.013$ for OLAS); a
paired $t$-test confirms this difference is not statistically
significant ($p=0.29$), nor are the gaps to ES+RLA and ES+OLA on
this dataset ($p=0.62$; see Appendix~C, Table~\ref{tab.significance}). On
all other datasets, OLAS outperforms all baselines by a substantial
margin. Among the baselines,
RS+OLA ranks second on several datasets, confirming that optimal
labeler assignment provides consistent benefit even when
combined with random sampling. MV achieves competitive
performance on several datasets by leveraging consensus labels,
but does so by labeling only Cap unique samples per cycle, which
reduces coverage of the unlabeled pool compared to other methods
under the same annotation budget. BADGE+RLA, which combines
uncertainty and diversity through pseudo-gradient embeddings,
performs competitively on Spambase ($0.915 \pm 0.008$) under
Noise Model~1, but shows degraded performance on the Ionosphere
dataset ($0.381 \pm 0.100$). This degradation occurs because the
pseudo-gradient embedding relies on raw input features rather
than learned neural representations, which can be less
informative for diversity-based selection on datasets where the
raw feature space does not align well with decision boundaries.
Under Noise Model~2, BADGE+RLA degrades more severely across
most datasets, highlighting that diversity-based sampling without
noise-aware labeler assignment is insufficient when label noise
levels are high.

\begin{table}[htbp]
  \centering
  \caption{F1 Score Results Based on Noise Model 1
           (mean $\pm$ standard deviation)}
  \resizebox{\columnwidth}{!}{%
  \begin{tabular}{c|c|c|c|c|c|c|c|c}
    \hline
    {Dataset} & RS+RLA & RS+OLA & ES+RLA & ES+OLA
             & {KC+RLA}
             & {MV}
             & {BADGE+RLA}
             & OLAS \\
    \hline
    Statlog (Heart)
      & $0.404 \pm 0.099$
      & $0.372 \pm 0.112$
      & $0.364 \pm 0.100$
      & $0.358 \pm 0.128$
      & {$0.376 \pm 0.116$}
      & {$0.309 \pm 0.135$}
      & {$0.366 \pm 0.106$}
      & $\mathbf{0.778 \pm 0.048}$ \\
    Ionosphere
      & $0.907 \pm 0.012$
      & $0.909 \pm 0.014$
      & $0.915 \pm 0.013$
      & $0.915 \pm 0.013$
      & {$\mathbf{0.918 \pm 0.012}$}
      & {$0.891 \pm 0.028$}
      & {$0.381 \pm 0.100$}
      & $0.916 \pm 0.013$ \\
    {Connectionist Bench (Sonar)}
      & $0.718 \pm 0.069$
      & $0.746 \pm 0.053$
      & $0.706 \pm 0.068$
      & $0.722 \pm 0.062$
      & {$0.712 \pm 0.070$}
      & {$0.711 \pm 0.054$}
      & {$0.698 \pm 0.071$}
      & $\mathbf{0.800 \pm 0.035}$ \\
    Spambase
      & $0.904 \pm 0.012$
      & $0.939 \pm 0.006$
      & $0.879 \pm 0.012$
      & $0.889 \pm 0.014$
      & {$0.896 \pm 0.013$}
      & {$0.923 \pm 0.008$}
      & {$0.915 \pm 0.008$}
      & $\mathbf{0.949 \pm 0.003}$ \\
    {Covertype}
      & {$0.396 \pm 0.019$}
      & {$0.409 \pm 0.021$}
      & {$0.388 \pm 0.021$}
      & {$0.408 \pm 0.030$}
      & {$0.398 \pm 0.029$}
      & {$0.438 \pm 0.030$}
      & {$0.354 \pm 0.021$}
      & {$\mathbf{0.452 \pm 0.020}$} \\
    \hline
  \end{tabular}%
  }
  \label{results.nm1}
\end{table}

{An important observation from
Table~\ref{results.nm1} is that RS+RLA outperforms ES+RLA across
all datasets. This highlights a key failure mode of
uncertainty-based sampling under label noise: entropy sampling
prioritizes high-uncertainty samples that are inherently more
difficult to label correctly. When these samples are assigned to
labelers at random, the probability of receiving a noisy label
is substantially higher, degrading downstream classifier
performance. Random sampling, by contrast, distributes the
labeling budget more uniformly across the difficulty spectrum,
reducing exposure to worst-case noise events when labeler
assignment is not optimized. This observation directly motivates
the OLAS design: sample selection and labeler assignment must be
jointly optimized to realize the benefits of uncertainty-based
active learning under label noise. Consistent with this, OLAS
combines uncertainty-based sampling with optimal labeler
assignment and outperforms both RS+RLA and ES+RLA across all
evaluated datasets and noise models.}

Table~\ref{results.nm2} shows the results for
{Noise Model~2. OLAS achieves the highest F1
score on all five datasets under this noise model, with
particularly large margins on the Ionosphere and Spambase
datasets. The consistency of OLAS performance across both noise
models provides strong evidence that the framework is robust to
the specific functional form of the noise model. Notably, under
Noise Model~2 the performance of all baselines on the Covertype
dataset drops substantially, while OLAS maintains a comparatively
strong result ($0.376 \pm 0.032$), demonstrating the importance
of joint optimization under more severe noise conditions.
BADGE+RLA collapses under Noise Model~2 on most datasets,
further confirming that diversity-based sampling without
noise-aware assignment is insufficient in high-noise settings.}

\begin{table}[htbp]
  \centering
  \caption{F1 Score Results Based on Noise Model 2
           (mean $\pm$ standard deviation)}
  \resizebox{\columnwidth}{!}{%
  \begin{tabular}{c|c|c|c|c|c|c|c|c}
    \hline
    {Dataset} & RS+RLA & RS+OLA & ES+RLA & ES+OLA
             & {KC+RLA}
             & {MV}
             & {BADGE+RLA}
             & OLAS \\
    \hline
    Statlog (Heart)
      & $0.230 \pm 0.052$
      & $0.211 \pm 0.047$
      & $0.205 \pm 0.044$
      & $0.205 \pm 0.045$
      & {$0.235 \pm 0.048$}
      & {$0.326 \pm 0.118$}
      & {$0.265 \pm 0.082$}
      & $\mathbf{0.707 \pm 0.104}$ \\
    Ionosphere
      & $0.099 \pm 0.039$
      & $0.093 \pm 0.028$
      & $0.108 \pm 0.033$
      & $0.107 \pm 0.031$
      & {$0.140 \pm 0.052$}
      & {$0.144 \pm 0.075$}
      & {$0.129 \pm 0.050$}
      & $\mathbf{0.913 \pm 0.014}$ \\
    {Connectionist Bench (Sonar)}
      & $0.302 \pm 0.131$
      & $0.230 \pm 0.134$
      & $0.179 \pm 0.103$
      & $0.162 \pm 0.091$
      & {$0.286 \pm 0.127$}
      & {$0.190 \pm 0.139$}
      & {$0.130 \pm 0.072$}
      & $\mathbf{0.747 \pm 0.042}$ \\
    Spambase
      & $0.230 \pm 0.025$
      & $0.230 \pm 0.076$
      & $0.143 \pm 0.017$
      & $0.154 \pm 0.027$
      & {$0.219 \pm 0.022$}
      & {$0.087 \pm 0.016$}
      & {$0.382 \pm 0.030$}
      & $\mathbf{0.949 \pm 0.003}$ \\
    {Covertype}
      & {$0.006 \pm 0.001$}
      & {$0.006 \pm 0.002$}
      & {$0.009 \pm 0.005$}
      & {$0.009 \pm 0.004$}
      & {$0.005 \pm 0.002$}
      & {$0.105 \pm 0.038$}
      & {$0.006 \pm 0.002$}
      & {$\mathbf{0.376 \pm 0.032}$} \\
    \hline
  \end{tabular}%
  }
  \label{results.nm2}
\end{table}

The other important observation is that OLA is useful when ES is
utilized for sampling. When RS is utilized, OLA does not improve
performance compared to randomly assigning samples to labelers.
This occurs because ES selects samples with the highest
uncertainty, making them more likely to receive noisy labels.
Thus, the use of OLA is beneficial in this context. Conversely,
the diversity of sample entropy is typically low when using RS,
meaning that the way samples are assigned to labelers does not
significantly impact the control of label noise. On the other
hand, ES+RLA performs worse than RS+RLA.

KC+RLA performs competitively on
several datasets, particularly under Noise Model~1, confirming
that diversity-based selection can be beneficial in low-noise
settings. However, KC+RLA does not consistently outperform OLAS,
and its purely diversity-driven selection without noise awareness
leads to substantially degraded performance under Noise Model~2,
where label noise levels are higher. Similarly, BADGE+RLA, which
combines uncertainty and diversity through pseudo-gradient
embeddings, shows competitive performance on some datasets under
Noise Model~1 but degrades significantly under Noise Model~2.
This confirms that diversity-based sampling alone, whether
geometric as in KC+RLA or gradient-based as in BADGE+RLA, is
insufficient: effective noise control requires the joint
optimization of sample selection and labeler assignment as
embodied in OLAS.

{Having established this accuracy advantage, we
next examine whether it comes at a disproportionate computational
cost, and whether it persists as dataset size grows.}

\subsection{Scalability and Computational Cost
Analysis} \label{sec.scalability}

{
This subsection reports the computational cost of all eight
methods across datasets of varying sizes, complementing the
accuracy comparison in Section~\ref{sec.benchmark}. Table~\ref{timing.table}
reports the average wall-clock time per AL cycle in seconds for
each method and dataset under both noise models. Each reported
time covers the full cycle, including sample selection, entropy
computation, labeler assignment, model retraining, and pool
update.}

\begin{table}[htbp]
  \centering
  \caption{{Average Computation Time per AL
  Cycle (seconds)}}
  \resizebox{\columnwidth}{!}{%
  \begin{tabular}{c|c|c|c|c|c|c|c|c|c}
    \hline
    {{Dataset}}
      & {Noise}
      & {RS+RLA}
      & {RS+OLA}
      & {ES+RLA}
      & {ES+OLA}
      & {KC+RLA}
      & {MV}
      & {BADGE+RLA}
      & {OLAS} \\
    \hline
    {Statlog (Heart)} & {NM1}
      & {$0.191$}
      & {$0.191$}
      & {$0.197$}
      & {$0.192$}
      & {$0.194$}
      & {$0.182$}
      & {$0.200$}
      & {$0.181$} \\
    {Statlog (Heart)} & {NM2}
      & {$0.188$}
      & {$0.181$}
      & {$0.190$}
      & {$0.193$}
      & {$0.198$}
      & {$0.187$}
      & {$0.182$}
      & {$0.173$} \\
    {Ionosphere} & {NM1}
      & {$0.208$}
      & {$0.205$}
      & {$0.216$}
      & {$0.216$}
      & {$0.238$}
      & {$0.189$}
      & {$0.190$}
      & {$0.209$} \\
    {Ionosphere} & {NM2}
      & {$0.212$}
      & {$0.213$}
      & {$0.219$}
      & {$0.219$}
      & {$0.243$}
      & {$0.191$}
      & {$0.188$}
      & {$0.209$} \\
    {Connectionist Bench (Sonar)}
      & {NM1}
      & {$0.200$}
      & {$0.197$}
      & {$0.212$}
      & {$0.215$}
      & {$0.208$}
      & {$0.187$}
      & {$0.194$}
      & {$0.204$} \\
    {Connectionist Bench (Sonar)}
      & {NM2}
      & {$0.208$}
      & {$0.208$}
      & {$0.213$}
      & {$0.209$}
      & {$0.209$}
      & {$0.185$}
      & {$0.190$}
      & {$0.194$} \\
    {Spambase} & {NM1}
      & {$0.845$}
      & {$0.790$}
      & {$0.948$}
      & {$0.934$}
      & {$1.135$}
      & {$0.245$}
      & {$0.541$}
      & {$0.745$} \\
    {Spambase} & {NM2}
      & {$0.963$}
      & {$0.912$}
      & {$0.960$}
      & {$0.939$}
      & {$1.250$}
      & {$0.243$}
      & {$0.584$}
      & {$0.752$} \\
    {Covertype} & {NM1}
      & {$0.302$}
      & {$0.376$}
      & {$0.311$}
      & {$0.415$}
      & {$6.838$}
      & {$0.145$}
      & {$21.269$}
      & {$0.492$} \\
    {Covertype} & {NM2}
      & {$0.314$}
      & {$0.404$}
      & {$0.325$}
      & {$0.420$}
      & {$6.820$}
      & {$0.166$}
      & {$20.691$}
      & {$0.440$} \\
    \hline
  \end{tabular}%
  }
  \label{timing.table}
\end{table}

{Several observations emerge from
Table~\ref{timing.table}. First, OLAS maintains competitive
per-cycle runtimes across all datasets, remaining within a
factor of two of the fastest methods on the UCI datasets. Second,
as dataset size increases from the small UCI datasets to Spambase
($4{,}601$ instances) and Covertype ($50{,}000$ instances), the
per-cycle time of OLAS scales gracefully, confirming that the
closed-form solution provided by Algorithm~\ref{alg} avoids the
combinatorial complexity of a brute-force search. Third, KC+RLA
exhibits high per-cycle time on both Spambase and Covertype,
reaching $6.84$ seconds per cycle on Covertype compared to $0.49$
seconds for OLAS, due to the quadratic nature of the greedy
$k$-center distance update step. Fourth, BADGE+RLA exhibits the
highest per-cycle time on Covertype, reaching $21.27$ seconds per
cycle under Noise Model~1. This is due to the cost of computing
pseudo-gradient embeddings of dimension $K \times d$ (where $K =
7$ classes and $d = 54$ features yields $378$-dimensional
embeddings) for each candidate sample, followed by $k$-means++
selection in that space. This substantial computational overhead,
combined with its weaker predictive performance relative to OLAS,
limits the practical applicability of BADGE+RLA at scale. Fifth,
MV achieves low per-cycle time on Spambase and Covertype because
it selects only Cap unique samples per cycle rather than the full
budget $B$, reducing the entropy evaluation cost. However, this
speed advantage comes at the cost of substantially lower F1
scores relative to OLAS across most settings, particularly under
Noise Model~2 where the gap reaches $0.769$ on the Ionosphere
dataset. These results confirm that OLAS is not only effective
but also computationally efficient, making it suitable for
large-scale annotation pipelines.}


\subsection{{Validation} on Crowdsourced Annotations}
\label{sec.music}

{
The experiments in Sections~\ref{sec.benchmark} and~\ref{sec.scalability} rely on
Noise Model~1 and Noise Model~2 to simulate label noise as a function of
labeler accuracy and sample entropy. While this simulation based approach
allows controlled and systematic comparison across methods, it does not by
itself demonstrate that OLAS performs well when labeler reliability and
label noise arise from real human annotators rather than an analytical
noise function. We therefore turn to a dataset with genuine crowdsourced
labels and evaluate OLAS and all comparison methods on the Music Genre
Classification dataset of \citet{rodrigues2013learning}, a publicly
available benchmark built from real crowdsourced annotations collected on
Amazon Mechanical Turk.}

The dataset consists of $700$ thirty second audio clips drawn from ten music
genres, each described by $124$ pre extracted tabular audio features,
including mel frequency cepstral coefficients and spectral features. Every
clip has a gold standard genre label established independently of the
crowd. In addition, $44$ Amazon Mechanical Turk workers each labeled a
subset of the clips, yielding $2{,}946$ individual annotations in total, an
average fill rate of $9.6$ percent of the full worker by sample annotation
matrix. Worker accuracy, measured against the gold labels, varies widely
across the crowd, with a mean of $0.733$ and a standard deviation of
$0.242$, reflecting the heterogeneous labeler reliability that motivates the
OLAS framework. A held out set of $300$ clips, disjoint from the $700$
annotated clips, is used as the test set for all reported results.
{Because the annotation matrix is only $9.6\%$
complete, worker-sample pairs that were not actually annotated are
filled in using the fitted noise model described in
Section~\ref{subsec.nm.beta} whenever a method queries them during
the active learning loop; we therefore treat this experiment as a
hybrid semi-synthetic evaluation, grounded in real crowdsourced
worker behavior but not a fully observed real-world annotation
process.}

\begin{figure}[t]
    \centering
    \includegraphics[scale=0.6]{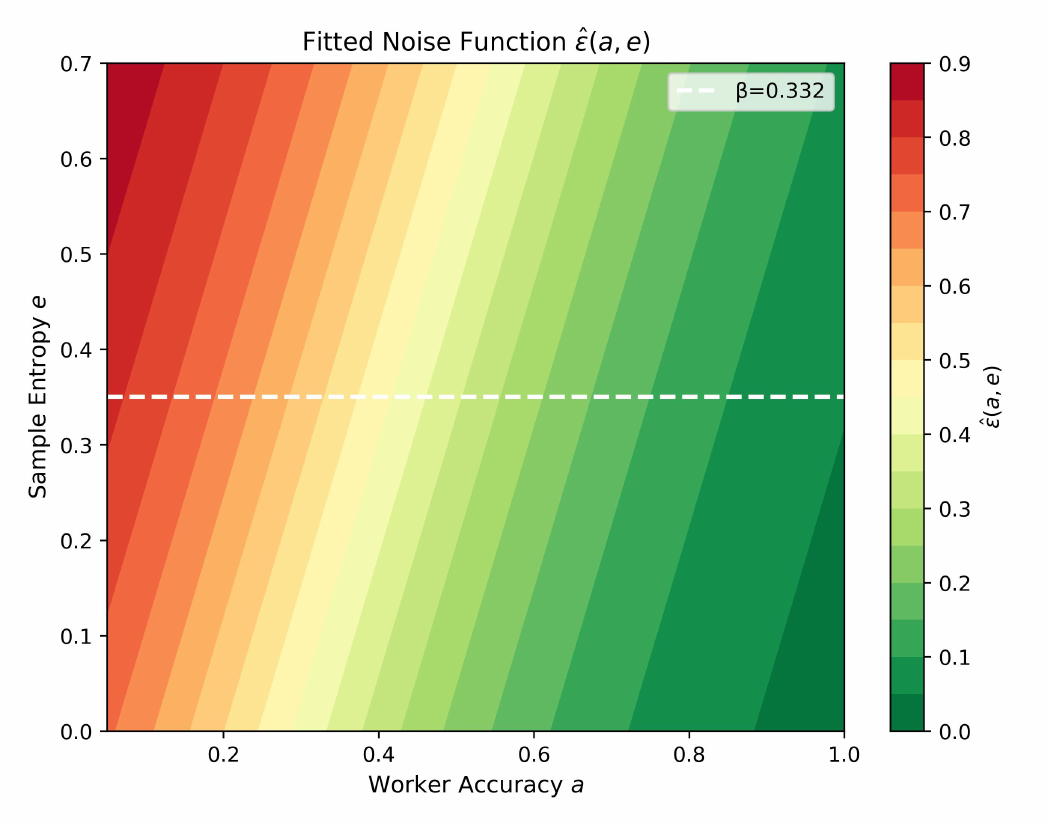}
    \caption{Validation of the Section~\ref{subsec.nm.beta} estimation pipeline on the Music
    dataset. The fitted noise function $\hat\epsilon(a, e)$, estimated from
    real annotator data on the gold held out subset, is shown with the
    estimated threshold $\beta = 0.332$ drawn as a dashed contour.}
    \label{fig.music.pipeline}
\end{figure}

To validate the estimation pipeline described in Section~\ref{subsec.nm.beta} directly on
real data, rather than only describing it, we set aside a gold labeled
subset of $140$ samples from the $700$ annotated clips and use the
remaining pool for the active learning experiment. The held out subset is
paired with the corresponding worker annotations to fit the logistic
regression noise function of Section~\ref{subsec.nm.beta}, using $589$ gold subset,
annotator pairs. The fitted model achieves a training accuracy of $0.806$
and preserves the monotonicity properties required by the theory, decreasing
in estimated labeler accuracy and increasing in sample entropy. The same
fitted model is used to estimate each of the $44$ workers' individual
accuracies directly from their observed labels; these estimates closely
track the true worker accuracies computed from the full annotation matrix,
with an estimated mean accuracy of $0.731$ against a true mean of $0.733$.
We similarly use the fitted pipeline to select $\beta$, obtaining
$\beta = 0.332$ as the $75$th percentile of the fitted noise distribution on
the gold subset, and use this estimated, data driven value of $\beta$
throughout the experiment below rather than an oracle or hand tuned choice.
Figure~\ref{fig.music.pipeline} shows the resulting fitted surface
$\hat\epsilon(a, e)$, estimated directly from real Music dataset annotator
data, together with the estimated threshold $\beta = 0.332$, drawn as a
dashed contour. The fitted surface is visibly decreasing in worker accuracy
and increasing in sample entropy across its full domain, confirming that
the monotonicity properties required by Section~\ref{subsec.nm.beta} hold when the pipeline
is estimated from real annotator data rather than assumed analytically.

For the active learning pool of the remaining $560$ samples, label outcomes
are realized using a hybrid procedure that combines the real annotation
structure with the estimated noise model: for each queried sample, labeler
pair, the estimated worker accuracy and the fitted noise function are used
to determine the probability that the assigned labeler's response is
corrected, and the observed label is then realized through a Bernoulli draw
against that probability, using the actual worker's response recorded in
the annotation matrix whenever it is available for that sample. This
mechanism allows us to evaluate all eight methods, including those that
require querying labelers who did not happen to annotate a given sample in
the original crowdsourcing collection, while still grounding every
probability in real, estimated worker behavior rather than a fully
synthetic noise function.

\begin{figure}[t]
    \centering
    \includegraphics[scale=0.45]{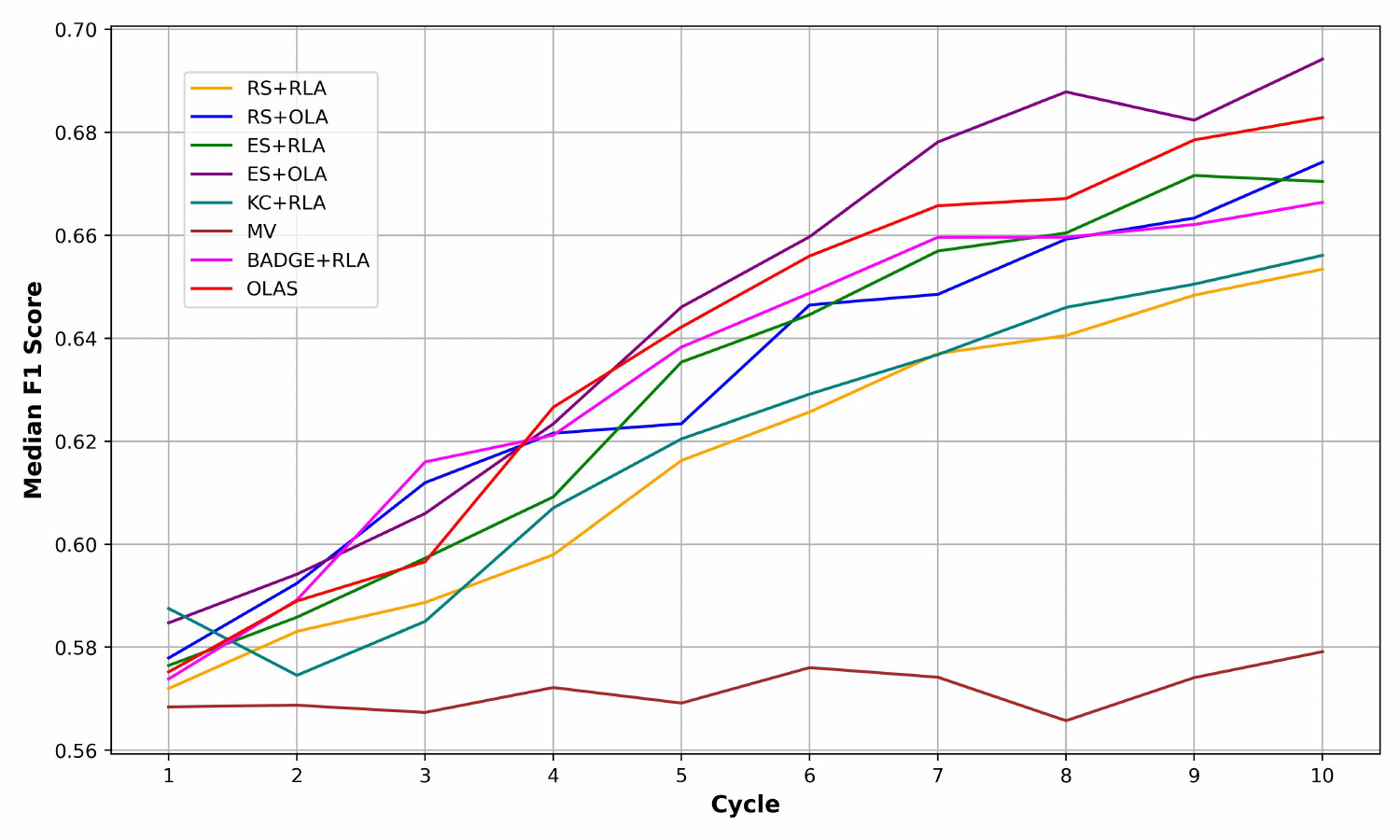}
    \caption{Median F1 score across 10 active learning cycles on the Music
    dataset, averaged over 30 replications.}
    \label{fig.music.median}
\end{figure}

\begin{figure}[t]
    \centering
    \includegraphics[scale=0.45]{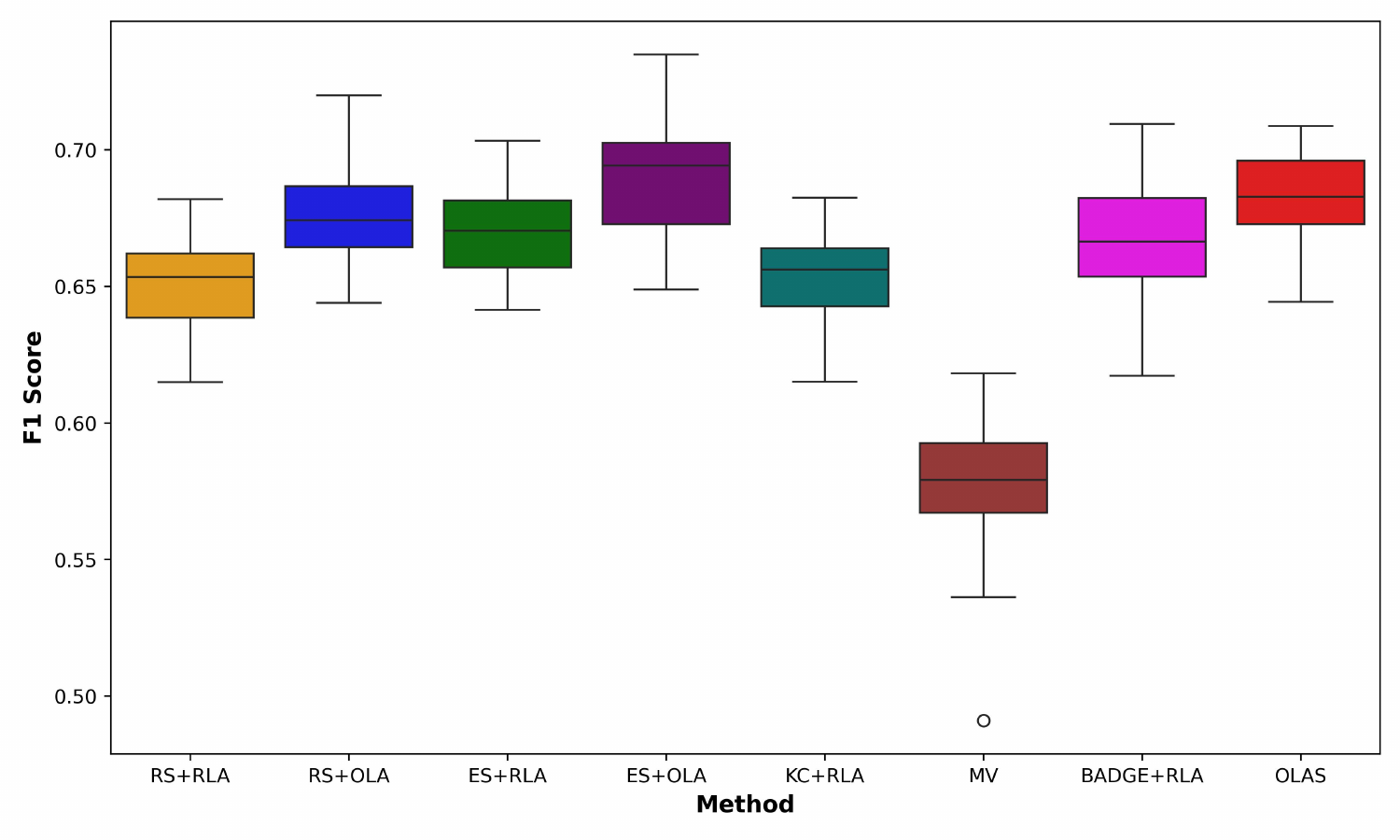}
    \caption{Box plot of final cycle (Cycle 10) macro F1 scores across 30
    replications on the Music dataset.}
    \label{fig.music.box}
\end{figure}

The experiment follows the same structure as Sections~\ref{sec.benchmark} and~\ref{sec.scalability}:
$T = 10$ active learning cycles, a query budget derived from the same formula used throughout the paper, and a {random forest} classifier trained
identically across all methods. Results are averaged over $30$
replications. Figure~\ref{fig.music.median} shows the median F1 score for
each method across all ten active learning cycles, and
Figure~\ref{fig.music.box} shows the distribution of final cycle F1 scores
across the $30$ replications.

At the final cycle, OLAS reaches a mean macro F1 of $0.682$ (standard
deviation $0.017$), close to the strongest method in the comparison,
ES+OLA, at $0.690$ (standard deviation $0.020$), a
difference of $0.008$ that a paired $t$-test confirms is
statistically significant ($p=0.008$) though modest in magnitude;
OLAS is not statistically distinguishable from RS+OLA ($p=0.11$) or
ES+RLA ($p=0.07$) on this dataset (Appendix~C, Table~\ref{tab.significance}). Both OLAS and ES+OLA clearly and consistently outperform
every method that does not use optimal labeler assignment: ES+RLA reaches
$0.671$, RS+OLA reaches $0.677$, BADGE+RLA reaches $0.666$, KC+RLA reaches
$0.654$, and RS+RLA reaches $0.650$. Majority voting, the consensus based
baseline, reaches only $0.577$, the weakest result among all eight methods.
This is consistent with the motivating premise of the paper: under sparse
and unevenly distributed real world annotation, of the kind observed in
this dataset where the average worker labels only a small fraction of the
pool, naive consensus labeling is not a reliable strategy, whereas methods
that explicitly account for labeler reliability, whether through random or
optimal assignment, are substantially more robust.

We further investigated, as an internal robustness check, why OLAS trails
ES+OLA by this small margin rather than exceeding it as it does in the
synthetic experiments of Section~\ref{sec.benchmark}. We verified that
OLAS's capacity constrained labeler assignment is not the explanation:
across all active learning cycles, the mean estimated noise, $\hat\epsilon$,
of the labelers OLAS actually assigns is statistically indistinguishable
from, and on average marginally lower than, the noise ES+OLA would incur by
assigning its own globally optimal labeler to the same sample ($0.111$
versus $0.112$ on average), and the capacity constraint in OLAS's
assignment step is never the binding factor that prevents it from reaching
its locally optimal labeler choice. We also confirmed that widening OLAS's
candidate sample window and varying the noise threshold $\beta$ over a
broad range of values does not close the small gap to ES+OLA, which rules
out those two design choices as the source of the difference. Taken
together, this evidence indicates that on a dataset with this particular
structure, namely wide variance in worker accuracy combined with a
classifier that already tolerates moderate label noise reasonably well,
entropy based sample selection paired with optimal labeler assignment alone
captures nearly all of the benefit that OLAS's joint optimization provides.
We view this as a genuine and informative characterization of OLAS's
operating regime rather than a limitation to be understated: the results
suggest that OLAS's joint optimization of sample and labeler is likely to
contribute the most additional value under tighter query budgets, more
constrained labeler pools, or noise structures further from the ones
realized in this dataset.

\subsection{Case Study: Automotive Warranty Claims} \label{sec.casestudy}

{Having validated OLAS on real crowdsourced
annotations in Section~\ref{sec.music}, we now turn to a second
real-world setting in which the labelers are domain experts rather
than crowd workers, and the labeling task is embedded in an
industrial production workflow.} In this case study, we address
the ACM problem, where daily warranty claims must be categorized
into appropriate classes to assist with root-cause identification.
We use warranty claims data from Ford Motor Company. Due to confidentiality concerns,
the data is not publicly available, and a complete description
cannot be provided. The dataset includes categorical and
numerical features explaining the characteristics of the vehicle
given in a warranty claim as well as the details and description
of the warranty issue. Each warranty claim can be categorized
into $36$ predefined classes. Finding the appropriate label for
a given warranty requires manual labeling, relying on expert
technicians who label the claims in their spare time. Due to
time limitations, the team is not able to label all claims.
Therefore, an AL framework is used to identify a subset of
claims for labeling in each cycle and to sequentially update the
model. Each technician may have different experience and skill
levels, which results in different labeling accuracy and may
lead to incorrect annotations during the labeling process.

Due to the lack of sufficient historical labeled data with
corresponding ground truth, we simulate labelers with varying
levels of accuracy. We employ Noise Model~1, as described in the
previous section, to generate label noise. {The
experiment is conducted over $30$ replications, each consisting
of $10$ AL cycles. The sampling budget is set to $B = 1994$
samples per cycle, with $M = 46$ labelers each having a capacity
of $\text{Cap} = 44$, derived from the training set size using
the same formula applied to the UCI datasets.} All eight methods are trained on a {random forest} classifier with identical settings in each replication, starting with $16$
percent of the training set as the initially labeled dataset.

\begin{figure}
\centering
\includegraphics[scale=0.45]{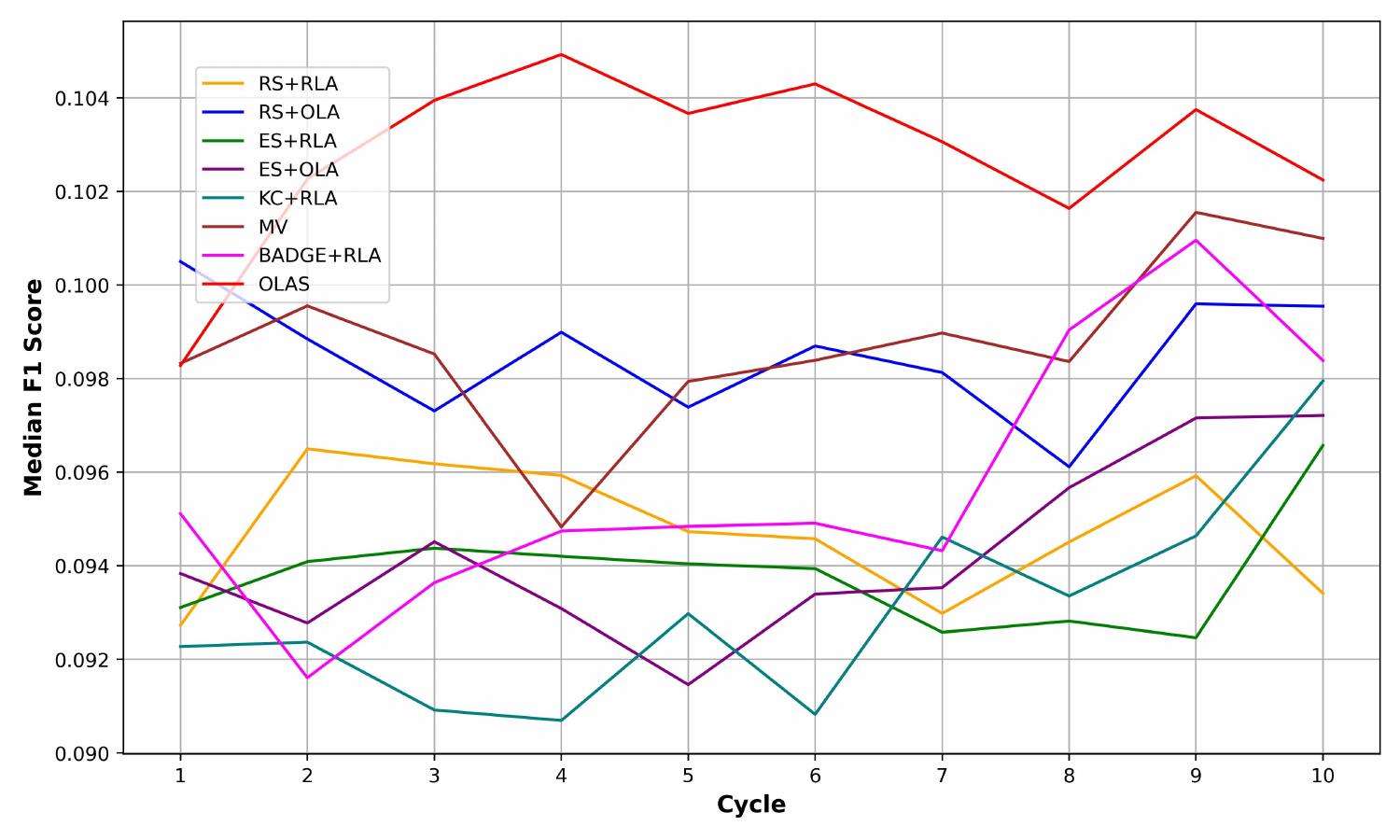}
\caption{{Median F1 score over 30 replications
across 10 active learning cycles on the Ford warranty claim
dataset. OLAS consistently achieves the highest median F1 score
across all cycles.}}
\label{fig:median}
\end{figure}

Figure~\ref{fig:median} displays the median F1 score for each
approach across $10$ cycles. {It is observed
that OLAS consistently achieves the highest median F1 score
across all cycles, demonstrating its ability to train a robust
model resistant to label noise. Among the baselines, MV achieves
competitive second-place performance by leveraging consensus
labels from all available labelers, though it labels only Cap
unique samples per cycle compared to the full budget $B$ labeled
by OLAS. RS+OLA and KC+RLA track closely behind MV, while
BADGE+RLA occupies the middle performance band, outperforming
RS+RLA, ES+RLA, and ES+OLA but falling short of OLAS, MV, and
RS+OLA throughout most cycles. The consistent separation between
OLAS and all baselines across all 10 cycles demonstrates that
the joint optimization of sample selection and labeler assignment
provides sustained benefits over the entire active learning
process.}

\begin{figure}
\centering
\includegraphics[scale=0.45]{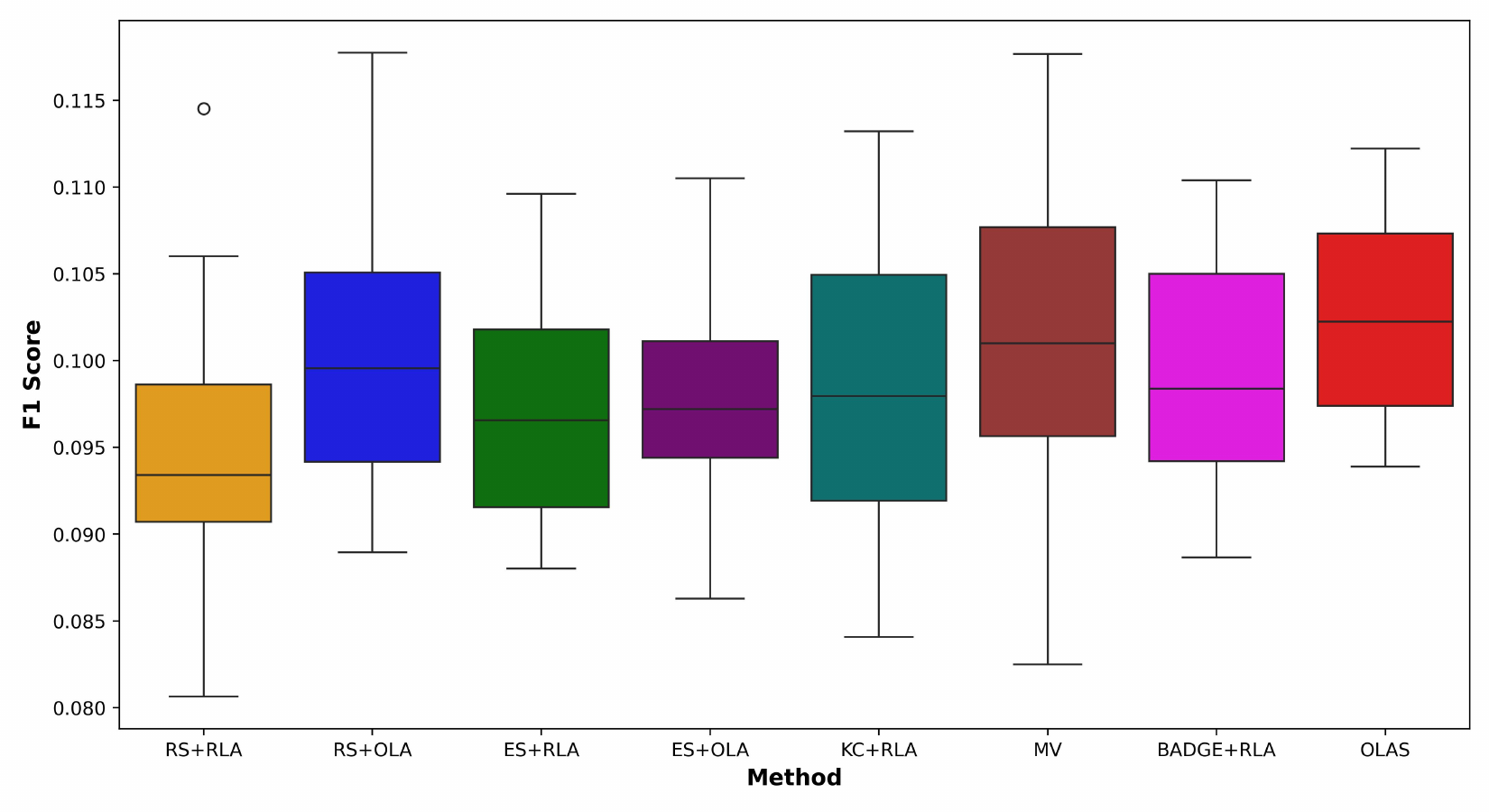}
\caption{{Box plot of F1 scores over 30
replications for the classifier at the final active learning
cycle on the Ford warranty claim dataset. OLAS achieves the
highest median with low variability across replications.}}
\label{fig:box}
\end{figure}

To better interpret the results, we focus on the box plot in
Figure~\ref{fig:box}, which presents all $30$ replications for
the final classifier trained at the end of the last cycle.
{The results indicate that OLAS achieves the
highest median F1 score at the final cycle, outperforming all
baseline methods. The variation of F1 scores obtained by OLAS
is among the smallest of all methods, indicating greater
consistency in outcomes. The mean $\pm$ standard deviation for
all methods at the final cycle is as follows: RS+RLA $0.094 \pm
0.008$, RS+OLA $0.101 \pm 0.008$, ES+RLA $0.097 \pm 0.006$,
ES+OLA $0.098 \pm 0.006$, KC+RLA $0.099 \pm 0.009$, MV $0.101
\pm 0.009$, BADGE+RLA $0.099 \pm 0.006$, and OLAS $0.103 \pm
0.006$. These results demonstrate that OLAS not only achieves
the highest mean F1 score but also maintains low variance,
confirming the stability and effectiveness of the proposed joint
optimization framework in a real-world industrial setting.
Notably, while MV achieves a mean F1 of $0.101$ which is close
to OLAS, it does so by labeling only Cap unique samples per
cycle. A paired $t$-test confirms neither this gap
nor the smaller gap to RS+OLA is statistically significant
($p=0.33$ and $p=0.32$, respectively; Appendix~C, Table~\ref{tab.significance}). OLAS matches or exceeds this performance while labeling
the full budget of unique samples, making it substantially more
practical for large-scale annotation pipelines where broad
coverage of the unlabeled pool is important. BADGE+RLA achieves
a mean F1 of $0.099$, comparable to KC+RLA and MV, but with
lower variance than MV, reflecting that its combined
uncertainty-diversity selection provides some benefit over purely
random assignment even without noise-aware labeler matching.}

\section{Conclusion}
\label{sec.conclusion}

In practice, inaccurate annotation from noisy oracles leads to inaccurate 
training of machine-learning models. In an active learning framework, highly 
uncertain data points are desirable choices for annotation. However, these 
points are typically associated with higher label noise. To control noise in 
the active learning framework, we study the operations within the active 
learning procedure. {We consider a noise function that depends 
on labeler accuracy and sample uncertainty, as quantified by the model's 
predictive entropy.} It is worth mentioning that accurate estimation of sample 
uncertainty and labeler accuracy plays an important role in selecting the best 
samples and the optimal assignment of samples to labelers. In practice, 
measuring the accuracies of labelers might be troublesome, requiring both 
budget and time. However, our results demonstrate that achieving accurate 
assessments for labelers is crucial and can lead to significant improvements 
in model performance.

Our experiments show that labeler assignment affects the generated noise, and 
hence, using an optimal assignment policy can be effective in reducing the 
number of incorrect labels. On the other hand, query selection is another 
factor that affects label noise. A formulation was proposed to optimally sample 
data points at each cycle of AL by maximizing information gain while bounding 
the label noise. Experimental results demonstrate the effectiveness of our 
sampling approach along with the optimal assignment policy in building a robust 
classification model within an AL framework. {We note that our 
framework focuses on the single-label-per-sample regime, which is appropriate 
for high-cost or time-sensitive annotation settings. Extending the model to 
incorporate redundant labeling and aggregation strategies under relaxed budget 
constraints is an interesting direction for future work (see Section~\ref{sec.methodology}).

{In our synthetic experiments (Section~\ref{sec.benchmark}), the corruption threshold $\alpha$ and OLAS's acceptance threshold $\beta$ are both set to $0.2$, reflecting a deployment where the evaluator's and the deployer's noise tolerance coincide. We note that this coupling means OLAS's acceptance rule is not fully independent of the corruption rule; decoupling the two in future work, for example by drawing corruption from a process independent of $\beta$, would further strengthen the comparison.}

{We also clarify the nature of our evaluation on the Music Genre Classification dataset in Section~\ref{sec.music}. While it is grounded in real crowdsourced annotations, the underlying annotation matrix is only $9.6\%$ complete, and unobserved worker-sample pairs are filled in using the fitted noise model rather than genuine worker responses. We therefore describe this experiment as a hybrid semi-synthetic evaluation: real crowdsourced behavior informs the noise model, but a majority of queried labels during the active learning loop are model-generated rather than directly observed. A denser real annotation matrix, or a protocol restricted to only observed worker-sample pairs, would strengthen this validation and is a natural direction for future work.}

{We also note that our noise pipeline (Section~\ref{subsec.nm.beta}) is validated using training accuracy and a visual monotonicity check rather than held-out calibration metrics or an enforced monotonicity constraint; a more rigorous treatment is left to future work.}

Although OLAS is presented for pool-based classification with noisy and heterogeneous labelers, the same optimization structure can be useful in settings where sample selection and resource assignment must be decided jointly. The method is most appropriate when labeler reliability varies across annotators, labeling costs are non-negligible, and uncertainty scores are available. Such settings include medical image annotation, legal document review, industrial quality inspection, and scientific data curation, where explicitly accounting for labeler reliability can reduce the risk of assigning highly uncertain samples to low-quality annotators.}


{
Beyond its operational benefits, systematically assigning samples to labelers based on estimated reliability raises ethical considerations that deserve explicit discussion. First, OLAS's assignment policy, by design, routes the most uncertain samples to the most accurate labelers, which means less accurate labelers are systematically assigned easier, lower-entropy samples. Over repeated cycles, this creates differential exposure by construction: some labelers are never given the opportunity to demonstrate competence on difficult cases, while others are consistently relied upon for the hardest decisions. In settings where labelers are also being evaluated, trained, or compensated based on the work they are assigned, this asymmetry could affect how labeler skill is perceived or rewarded independently of the quality of their actual judgment.

Second, because OLAS acquires only a single label per sample, there is no redundancy to catch or correct a mistaken assignment. If the estimated noise function or labeler accuracy is wrong for a given sample, whether due to poor estimation, a labeler's reliability changing over time, or an atypical instance that does not fit the labeler's usual area of competence, that error propagates directly into the training set with no opportunity for another labeler's judgment to offset it. This risk is more consequential in high-stakes domains such as medical annotation or safety-critical industrial inspection, where the cost of a single incorrect label reaching the training set, or downstream model, can be substantial and is not mitigated by the consensus mechanisms available to multi-label approaches.

Third, the systematic and uneven distribution of easy and hard cases across labelers raises a fairness concern among the human annotators themselves, distinct from the fairness of the resulting classifier. A labeler who is judged, formally or informally, on the difficulty of the cases they are trusted with may be disadvantaged simply by having a lower estimated accuracy early in the process, regardless of whether that estimate reflects a stable underlying skill difference or an artifact of limited early data. Because OLAS's assignment decisions compound over active learning cycles, an early, possibly noisy, reliability estimate can influence which labelers continue to receive difficult, and potentially more visible or valued, work throughout the remainder of the labeling process.

These concerns do not have a complete algorithmic solution within the scope of this paper, but two properties of the framework and its deployment can partially mitigate them. First, because labeler accuracy estimates are updated sequentially as more labeled data become available (Section~\ref{subsec.nm.beta}), an early and possibly noisy estimate is not permanent: as evidence accumulates, a labeler's estimated reliability, and therefore the difficulty of the samples routed to them, can shift accordingly, which limits how long any single early estimate can dictate a labeler's assignment history. Second, since OLAS produces an explicit, auditable record of which sample was assigned to which labeler and why, a deployer can monitor the distribution of sample difficulty across labelers over time and intervene, for example by periodically reserving a small share of the budget for samples outside a labeler's usual assignment profile, without altering the underlying optimization. We view these as practical safeguards for responsible deployment rather than a resolution of the underlying tension, and we see a more formal treatment of fairness-aware labeler assignment as a valuable direction for future work.
}








\bibliography{ref}

\newpage


\setcounter{theorem}{0}
\section*{Appendix A}

In this appendix, we prove Theorem~\ref{thm.mm} from Section~\ref{subsec.labeler.assign}.

{
\subsection*{Restatement of Theorem~1}

\medskip\noindent\textbf{Theorem 1 (Optimal Assignment Policy).} {\itshape
Assume without loss of generality that $e_1 \geq e_2 \geq \cdots \geq e_{N_t}$ 
and $a_1 \geq a_2 \geq \cdots \geq a_M$. The optimal solution to 
Model~\eqref{mm.model} is
\begin{equation*}
z_{1j}^* =
\begin{cases}
1, & \text{if } j \leq c_1, \\
0, & \text{otherwise},
\end{cases}
\quad \forall j \in \{1, \ldots, N_t\},
\end{equation*}
and
\begin{equation*}
z_{ij}^* =
\begin{cases}
1, & \text{if } \sum_{k=1}^{i-1} c_k < j \leq \sum_{k=1}^{i} c_k, \\
0, & \text{otherwise},
\end{cases}
\end{equation*}
for all $i \in \{2, \ldots, M\}$ and $j \in \{1, \ldots, N_t\}$.}
\medskip
}

\noindent\textit{Proof.}

For any $j \in [N_t]$, there exists exactly one $i \in [M]$ such that 
$\sum_{k=1}^{i-1} c_k < j \leq \sum_{k=1}^i c_k$. Therefore, 
equations~\eqref{sol1} and \eqref{sol2} imply that $\sum_{i=1}^{M} z^*_{ij} = 1$. 
Additionally, since $N_t \leq \sum_{k=1}^{M} c_k$ and there are at most $c_k$ 
values of $j \in [N_t]$ such that $\sum_{k=1}^{i-1} c_k < j \leq \sum_{k=1}^i c_k$, 
we have $\sum_{j=1}^{N_t} z_{ij} \leq c_i, \ \forall i \in [M]$. Hence, 
equations~\eqref{sol1} and \eqref{sol2} define a feasible solution $z^*$ for 
Model~\eqref{mm.model}. To prove the optimality of $z^*$, we define the set
\begin{align*}
    \Pi^* = \{(i,j) \mid i\in [M],\ j\in [N_t],\ z^*_{ij} = 1\}.
\end{align*}
For a given feasible solution $z$, we define
\begin{align*}
    \Pi = \{(i,j) \mid i\in [M],\ j\in [N_t],\ z_{ij} = 1\}.
\end{align*}
We define the objective function as $f(z) = \max_{j} \sum_{i=1}^{M} \epsilon(a_i, e_j) z_{ij}$, 
and since $\sum_{i=1}^{M} z_{ij} = 1, \forall j \in [N_t]$, for any feasible 
solution $z$, there exists $(i,j) \in \Pi$ such that $f(z) = \epsilon(a_i, e_j)$. 
Let us assume $f(z^*) = \epsilon(a_{i^*}, e_{j^*})$. We then consider the 
following cases:
\begin{enumerate}
    \item If $(i^*, j^*) \in \Pi$, then $f(z) \geq \epsilon(a_{i^*}, e_{j^*}) = f(z^*)$.
    \item If $(i^*, j^*) \notin \Pi$, there exists $w \in [M]$ such that 
    $(w, j^*) \in \Pi$ and $w \neq i^*$. In this case, there are two possibilities:
    \begin{enumerate}
        \item If $w > i^*$, then $f(z) \geq \epsilon(a_w, e_{j^*}) \geq 
        \epsilon(a_{i^*}, e_{j^*}) = f(z^*)$.
        \item If $w < i^*$, the remaining capacity for all labelers with an index 
        less than or equal to $i^*$ is $\sum_{k=1}^{i^*} c_k - 1$. On the other 
        hand, there are $j^*-1$ available points for assignment with an index less 
        than $j^*$. From equations~\eqref{sol1} and \eqref{sol2}, we can infer that
        \begin{align*}
            j^* \leq \sum_{k=1}^{i^*} c_k \implies j^*-1 \leq \sum_{k=1}^{i^*} c_k - 1.
        \end{align*}
        This means the number of available points for assignment with an index 
        less than $j^*$ is less than or equal to the remaining capacity for all 
        labelers with an index less than or equal to $i^*$. Therefore, there 
exists $u \in [N_t]$ and $v \in [M]$ such that $u < j^*$, $v \geq i^*$, 
and ${(v,u) \in \Pi}$. This implies
\begin{align*}
    f(z) \geq \epsilon(a_v, e_u) \geq \epsilon(a_{i^*}, e_{j^*}) = f(z^*).
\end{align*}
    \end{enumerate}
\end{enumerate}
Thus, for any feasible solution $z$, $f(z) \geq f(z^*)$, and $z^*$ is the 
optimal solution. \hfill$\square$

\newpage


\setcounter{theorem}{1}
\section*{Appendix B}

In this appendix, we prove Theorem~\ref{thm.qm} from Section~\ref{subsec.qm}.

{
\subsection*{Restatement of Theorem~2}

\medskip\noindent\textbf{Theorem 2 (Optimal Solution Characterization).} {\itshape
Assume without loss of generality that $e_1 \geq e_2 \geq \cdots \geq e_{u_t}$ 
and $a_1 \geq a_2 \geq \cdots \geq a_M$.

For each $i \in [M]$, if it exists, define $r_i$ as the smallest integer $r$ 
satisfying
\begin{equation*}
r_i = \arg\min_{r \in \mathbb{Z}} \left\{ r :
\begin{array}{l}
\epsilon(a_i, e_r) \leq \beta, \\
r_{i-1} + c_{i-1} \leq r, \\
1 \leq r \leq u_t
\end{array}
\right\}.
\end{equation*}
By convention, define $r_0 = 0$ and $c_0 = 0$. If $r_u$ does not exist for 
some $u \in [M]$, then $r_v$ does not exist for all $v \in \{u, \ldots, M\}$.

Then, the optimal solution to Model~\eqref{qm.model} is characterized by
\begin{equation*}
y_{ij}^* =
\begin{cases}
1, & \text{if $r_i$ exists and $r_i \leq j < r_i + c_i$,} \\
0, & \text{otherwise,}
\end{cases}
\end{equation*}
for all $i \in [M]$ and $j \in [u_t]$, and
\begin{equation*}
z_i^* =
\begin{cases}
1, & \text{if } c_i - \sum_{j=1}^{u_t} y_{ij}^* > 0, \\
0, & \text{otherwise},
\end{cases}
\quad \forall i \in [M].
\end{equation*}}
\medskip
}

\noindent\textit{Proof.}

The proof addresses both the feasibility and optimality of the 
solutions~\eqref{qm.sol.y} and \eqref{qm.sol.z}.

\begin{itemize}
    \item \textbf{Feasibility:} We show that $y^*$ and $z^*$, defined 
    in~\eqref{qm.sol.y} and \eqref{qm.sol.z}, satisfy all constraints.

    \begin{itemize}
        \item \textit{Constraint~\eqref{qm.c1}:} For each $j \in [u_t]$, if 
        there exists $r_i$ such that $r_i \leq j < r_i + c_i$, then 
        $\not\exists\, i' \in [M]$ with $i' \neq i$ and 
        $r_{i'} \leq j < r_{i'} + c_{i'}$; hence $\sum_{i=1}^{M} y^*_{ij} = 1$. 
        If no such $i$ exists, then $\sum_{i=1}^{M} y^*_{ij} = 0$. In general, 
        $\sum_{i=1}^{M} y^*_{ij} \leq 1$.

        \item \textit{Constraint~\eqref{qm.c2}:} For each $i \in [M]$, if $r_i$ 
        does not exist, then $\sum_{j=1}^{u_t} y^*_{ij} = 0 \leq c_i$. If $r_i$ 
        exists, there are at most $c_i$ values of $j$ such that 
        $r_i \leq j < r_i + c_i$, so $\sum_{j=1}^{u_t} y^*_{ij} \leq c_i$.

        \item \textit{Constraint~\eqref{qm.c3}:} For each $i \in [M]$, 
        $y^*_{ij} = 1$ only if $r_i$ exists. By definition of $r_i$, 
        $\epsilon(a_i, e_{r_i}) \leq \beta$. Since the noise function is 
        increasing in entropy and $e_j \leq e_{r_i}$ for all 
        $j \geq r_i$, we have $\epsilon(a_i, e_j) \leq \beta$ for all $j$ 
        such that $r_i \leq j < r_i + c_i$, ensuring the constraint holds.

        \item \textit{Constraint~\eqref{qm.c4}:} If $y^*_{ij} = 1$, then $r_i$ 
        exists and $r_i \leq j < r_i + c_i$. Suppose $y^*_{kl} = 1$ with $k < i$; 
        then by~\eqref{qm.sol.y}, $r_k \leq l < r_k + c_k$. From the definition 
        of $r_i$,
        \begin{align*}
            r_k \leq l < r_k + c_k \leq r_i \leq j < r_i + c_i.
        \end{align*}
        Thus $l < j$, confirming that no data point with index greater than $j$ 
        is assigned to a labeler with index smaller than $i$.

        \item \textit{Constraints~\eqref{qm.c5}--\eqref{qm.c7}:} Since 
        Constraint~\eqref{qm.c2} holds, $\sum_{j=1}^{u_t} y^*_{ij} \leq c_i$ 
        for all $i \in [M]$.
        \begin{itemize}
            \item If $\sum_{j=1}^{u_t} y^*_{ij} = c_i$, then 
            Constraints~\eqref{qm.c5} and \eqref{qm.c6} imply $z_i^* = 0$, and 
            Constraint~\eqref{qm.c7} reduces to 
            $\sum_{j=1}^{u_t} y^*_{i+1,j} \leq c_{i+1}$, which holds 
            by Constraint~\eqref{qm.c2}.
            \item If $\sum_{j=1}^{u_t} y^*_{ij} < c_i$, then 
            Constraints~\eqref{qm.c5} and \eqref{qm.c6} imply $z_i^* = 1$, and 
            Constraint~\eqref{qm.c7} requires 
            $\sum_{j=1}^{u_t} y^*_{i+1,j} = 0$. This holds because: if $r_i$ 
            does not exist, then $r_{i+1}$ does not exist; if $r_i$ exists, then 
            $\sum_{j=1}^{u_t} y^*_{ij} < c_i$ implies $r_i \leq u_t < r_i + c_i$, 
            which means $r_{i+1}$ does not exist. In both cases, 
            $\sum_{j=1}^{u_t} y^*_{i+1,j} = 0$.
        \end{itemize}
    \end{itemize}

    Hence, $y^*$ and $z^*$ satisfy all constraints in Model~\eqref{qm.model} 
    and constitute a feasible solution.

    \item \textbf{Optimality:} For any feasible solution $(y, z)$, we show that
    \begin{equation*}
        \sum_{j=1}^{u_t}\sum_{i=1}^{M} e_j y^*_{ij} 
        \geq \sum_{j=1}^{u_t}\sum_{i=1}^{M} e_j y_{ij}.
    \end{equation*}
    Define
    \begin{align*}
        L^* &= \left\{j \in [u_t] \;\middle|\; \sum_{i=1}^{M} y^*_{ij} = 1\right\},\\
        L   &= \left\{j \in [u_t] \;\middle|\; \sum_{i=1}^{M} y_{ij}  = 1\right\},
    \end{align*}
    and let $\bar{r} = \max\left\{j \in [u_t] \mid j \in L^*\right\}$ denote 
    the index of the last sample selected in the optimal solution.

    We first show that for all $j \leq \bar{r}$, if $\sum_{i=1}^M y_{ij} = 1$ 
    for any feasible solution $y$, then $\sum_{i=1}^M y^*_{ij} = 1$. If 
    $\sum_{i=1}^M y_{ij} = 1$, there exists $i' \in [M]$ such that 
    $\epsilon(a_{i'}, e_j) \leq \beta$. This implies there exists $i \leq i'$ 
    such that $r_i$ exists and $y^*_{ij} = 1$, i.e., 
    $\sum_{i=1}^{M} y^*_{ij} = 1$. Thus, any sample labeled in a feasible 
    solution with index $j \leq \bar{r}$ is also labeled in the optimal solution.

  Let $C = \left\{j \leq \bar{r} \;\middle|\; {\sum_{i=1}^{M} y_{ij} = 1}\right\}$ and 
$R = \left\{j > \bar{r} \;\middle|\; {\sum_{i=1}^{M} y_{ij} = 1}\right\}$, so $L = C \cup R$. From the 
    statement above, $C \subseteq L^*$. Letting $R^* = \{j \in L^* \mid j \notin C\}$, 
    we have $L^* = C \cup R^*$. Therefore,
    \begin{align*}
        f^* - f 
        &= \sum_{j \in R^*} e_j - \sum_{j \in R} e_j.
    \end{align*}
    Since $j \leq \bar{r}$ for all $j \in R^*$ and $j > \bar{r}$ for all 
    $j \in R$, and the entropy sequence is non-increasing, we have 
    $e_j \geq e_{j'}$ for all $j \in R^*$ and $j' \in R$. Hence $f^* \geq f$, 
    and the optimal solution is established.

\end{itemize}
\hfill$\square$

%
%
%

\newpage

\section*{Appendix C}
\label{app.significance}

For every dataset, noise model, and pair of methods reported in
Section~\ref{sec.experiments}, we compared OLAS to each baseline using a
paired $t$-test and a Wilcoxon signed-rank test, matched by replication
index. Pairing is valid for the seven core methods (RS+RLA, RS+OLA, ES+RLA,
ES+OLA, OLAS, MV, KC+RLA) and, for the Music dataset, BADGE+RLA, since these
were computed within the same per-replication run and therefore share the
same initial labeled set, labeler accuracy pool, and train/test split for a
given replication index. BADGE+RLA on the five UCI/Covertype datasets and on
Ford (marked $^\dagger$ below) was run as a separate, later execution and
does not share this per-replication randomness; those comparisons are
therefore reported using an unpaired Welch's $t$-test and a Mann--Whitney
$U$ test instead, and no paired confidence interval is reported for them.
Table~\ref{tab.significance} reports, for each comparison, the mean
difference in F1 (OLAS minus baseline), the 95\% confidence interval where
applicable, the $t$-test and Wilcoxon/Mann--Whitney $p$-values, and whether
the difference is significant at $\alpha=0.05$.

{
\footnotesize
\begin{longtable}{p{3.3cm}p{1.6cm}p{1.5cm}p{2.3cm}p{1.5cm}p{1.5cm}p{0.9cm}}
\caption{Paired (or unpaired$^\dagger$, where noted) significance tests
comparing OLAS to each baseline method, by dataset and noise model. Mean
difference is OLAS minus the baseline; positive values favor OLAS.}
\label{tab.significance} \\
\hline
\textbf{Dataset (NM)} & \textbf{Baseline} & \textbf{Mean diff.} & \textbf{95\% CI} & \textbf{$p$ ($t$)} & \textbf{$p$ (W/MW)} & \textbf{Sig.} \\
\hline
\endfirsthead
\hline
\textbf{Dataset (NM)} & \textbf{Baseline} & \textbf{Mean diff.} & \textbf{95\% CI} & \textbf{$p$ ($t$)} & \textbf{$p$ (W/MW)} & \textbf{Sig.} \\
\hline
\endhead
\hline
\endfoot
Heart Statlog (NM1) & RS+RLA & +0.3738 & [0.353, 0.395] & $<0.001$ & $<0.001$ & Yes \\
Heart Statlog (NM1) & RS+OLA & +0.4060 & [0.383, 0.429] & $<0.001$ & $<0.001$ & Yes \\
Heart Statlog (NM1) & ES+RLA & +0.4140 & [0.393, 0.435] & $<0.001$ & $<0.001$ & Yes \\
Heart Statlog (NM1) & ES+OLA & +0.4198 & [0.394, 0.445] & $<0.001$ & $<0.001$ & Yes \\
Heart Statlog (NM1) & MV & +0.4691 & [0.442, 0.496] & $<0.001$ & $<0.001$ & Yes \\
Heart Statlog (NM1) & KC+RLA & +0.4016 & [0.378, 0.425] & $<0.001$ & $<0.001$ & Yes \\
Heart Statlog (NM1) & BADGE+RLA$^\dagger$ & +0.4114 & -- & $<0.001$ & $<0.001$ & Yes \\
Heart Statlog (NM2) & RS+RLA & +0.4768 & [0.454, 0.499] & $<0.001$ & $<0.001$ & Yes \\
Heart Statlog (NM2) & RS+OLA & +0.4956 & [0.474, 0.517] & $<0.001$ & $<0.001$ & Yes \\
Heart Statlog (NM2) & ES+RLA & +0.5020 & [0.479, 0.524] & $<0.001$ & $<0.001$ & Yes \\
Heart Statlog (NM2) & ES+OLA & +0.5023 & [0.480, 0.525] & $<0.001$ & $<0.001$ & Yes \\
Heart Statlog (NM2) & MV & +0.3807 & [0.353, 0.408] & $<0.001$ & $<0.001$ & Yes \\
Heart Statlog (NM2) & KC+RLA & +0.4718 & [0.449, 0.495] & $<0.001$ & $<0.001$ & Yes \\
Heart Statlog (NM2) & BADGE+RLA$^\dagger$ & +0.4417 & -- & $<0.001$ & $<0.001$ & Yes \\
Ionosphere (NM1) & RS+RLA & +0.0090 & [0.006, 0.012] & $<0.001$ & $<0.001$ & Yes \\
Ionosphere (NM1) & RS+OLA & +0.0068 & [0.003, 0.010] & $<0.001$ & $<0.001$ & Yes \\
Ionosphere (NM1) & ES+RLA & +0.0008 & [-0.002, 0.004] & 0.615 & 0.690 & No \\
Ionosphere (NM1) & ES+OLA & +0.0008 & [-0.002, 0.004] & 0.615 & 0.690 & No \\
Ionosphere (NM1) & MV & +0.0255 & [0.019, 0.032] & $<0.001$ & $<0.001$ & Yes \\
Ionosphere (NM1) & KC+RLA & -0.0018 & [-0.005, 0.002] & 0.292 & 0.268 & No \\
Ionosphere (NM1) & BADGE+RLA$^\dagger$ & +0.5349 & -- & $<0.001$ & $<0.001$ & Yes \\
Ionosphere (NM2) & RS+RLA & +0.8137 & [0.805, 0.822] & $<0.001$ & $<0.001$ & Yes \\
Ionosphere (NM2) & RS+OLA & +0.8193 & [0.813, 0.826] & $<0.001$ & $<0.001$ & Yes \\
Ionosphere (NM2) & ES+RLA & +0.8045 & [0.798, 0.811] & $<0.001$ & $<0.001$ & Yes \\
Ionosphere (NM2) & ES+OLA & +0.8056 & [0.799, 0.812] & $<0.001$ & $<0.001$ & Yes \\
Ionosphere (NM2) & MV & +0.7685 & [0.753, 0.784] & $<0.001$ & $<0.001$ & Yes \\
Ionosphere (NM2) & KC+RLA & +0.7728 & [0.762, 0.783] & $<0.001$ & $<0.001$ & Yes \\
Ionosphere (NM2) & BADGE+RLA$^\dagger$ & +0.7840 & -- & $<0.001$ & $<0.001$ & Yes \\
Sonar (NM1) & RS+RLA & +0.0818 & [0.066, 0.098] & $<0.001$ & $<0.001$ & Yes \\
Sonar (NM1) & RS+OLA & +0.0546 & [0.042, 0.067] & $<0.001$ & $<0.001$ & Yes \\
Sonar (NM1) & ES+RLA & +0.0948 & [0.079, 0.111] & $<0.001$ & $<0.001$ & Yes \\
Sonar (NM1) & ES+OLA & +0.0784 & [0.064, 0.093] & $<0.001$ & $<0.001$ & Yes \\
Sonar (NM1) & MV & +0.0892 & [0.077, 0.102] & $<0.001$ & $<0.001$ & Yes \\
Sonar (NM1) & KC+RLA & +0.0886 & [0.073, 0.104] & $<0.001$ & $<0.001$ & Yes \\
Sonar (NM1) & BADGE+RLA$^\dagger$ & +0.1023 & -- & $<0.001$ & $<0.001$ & Yes \\
Sonar (NM2) & RS+RLA & +0.4456 & [0.418, 0.473] & $<0.001$ & $<0.001$ & Yes \\
Sonar (NM2) & RS+OLA & +0.5167 & [0.491, 0.543] & $<0.001$ & $<0.001$ & Yes \\
Sonar (NM2) & ES+RLA & +0.5682 & [0.546, 0.590] & $<0.001$ & $<0.001$ & Yes \\
Sonar (NM2) & ES+OLA & +0.5856 & [0.565, 0.607] & $<0.001$ & $<0.001$ & Yes \\
Sonar (NM2) & MV & +0.5576 & [0.528, 0.587] & $<0.001$ & $<0.001$ & Yes \\
Sonar (NM2) & KC+RLA & +0.4609 & [0.435, 0.487] & $<0.001$ & $<0.001$ & Yes \\
Sonar (NM2) & BADGE+RLA$^\dagger$ & +0.6173 & -- & $<0.001$ & $<0.001$ & Yes \\
Spambase (NM1) & RS+RLA & +0.0449 & [0.043, 0.047] & $<0.001$ & $<0.001$ & Yes \\
Spambase (NM1) & RS+OLA & +0.0103 & [0.009, 0.011] & $<0.001$ & $<0.001$ & Yes \\
Spambase (NM1) & ES+RLA & +0.0693 & [0.067, 0.072] & $<0.001$ & $<0.001$ & Yes \\
Spambase (NM1) & ES+OLA & +0.0599 & [0.057, 0.063] & $<0.001$ & $<0.001$ & Yes \\
Spambase (NM1) & MV & +0.0258 & [0.024, 0.027] & $<0.001$ & $<0.001$ & Yes \\
Spambase (NM1) & KC+RLA & +0.0531 & [0.051, 0.056] & $<0.001$ & $<0.001$ & Yes \\
Spambase (NM1) & BADGE+RLA$^\dagger$ & +0.0335 & -- & $<0.001$ & $<0.001$ & Yes \\
Spambase (NM2) & RS+RLA & +0.7185 & [0.714, 0.723] & $<0.001$ & $<0.001$ & Yes \\
Spambase (NM2) & RS+OLA & +0.7190 & [0.704, 0.734] & $<0.001$ & $<0.001$ & Yes \\
Spambase (NM2) & ES+RLA & +0.8055 & [0.802, 0.809] & $<0.001$ & $<0.001$ & Yes \\
Spambase (NM2) & ES+OLA & +0.7941 & [0.789, 0.799] & $<0.001$ & $<0.001$ & Yes \\
Spambase (NM2) & MV & +0.8619 & [0.859, 0.865] & $<0.001$ & $<0.001$ & Yes \\
Spambase (NM2) & KC+RLA & +0.7297 & [0.725, 0.734] & $<0.001$ & $<0.001$ & Yes \\
Spambase (NM2) & BADGE+RLA$^\dagger$ & +0.5671 & -- & $<0.001$ & $<0.001$ & Yes \\
Covertype (NM1) & RS+RLA & +0.0558 & [0.049, 0.063] & $<0.001$ & $<0.001$ & Yes \\
Covertype (NM1) & RS+OLA & +0.0429 & [0.035, 0.050] & $<0.001$ & $<0.001$ & Yes \\
Covertype (NM1) & ES+RLA & +0.0642 & [0.056, 0.073] & $<0.001$ & $<0.001$ & Yes \\
Covertype (NM1) & ES+OLA & +0.0437 & [0.033, 0.054] & $<0.001$ & $<0.001$ & Yes \\
Covertype (NM1) & MV & +0.0134 & [0.004, 0.023] & 0.006 & 0.005 & Yes \\
Covertype (NM1) & KC+RLA & +0.0536 & [0.043, 0.064] & $<0.001$ & $<0.001$ & Yes \\
Covertype (NM1) & BADGE+RLA$^\dagger$ & +0.0982 & -- & $<0.001$ & $<0.001$ & Yes \\
Covertype (NM2) & RS+RLA & +0.3698 & [0.361, 0.379] & $<0.001$ & $<0.001$ & Yes \\
Covertype (NM2) & RS+OLA & +0.3701 & [0.361, 0.379] & $<0.001$ & $<0.001$ & Yes \\
Covertype (NM2) & ES+RLA & +0.3667 & [0.358, 0.376] & $<0.001$ & $<0.001$ & Yes \\
Covertype (NM2) & ES+OLA & +0.3666 & [0.358, 0.376] & $<0.001$ & $<0.001$ & Yes \\
Covertype (NM2) & MV & +0.2706 & [0.258, 0.283] & $<0.001$ & $<0.001$ & Yes \\
Covertype (NM2) & KC+RLA & +0.3706 & [0.361, 0.380] & $<0.001$ & $<0.001$ & Yes \\
Covertype (NM2) & BADGE+RLA$^\dagger$ & +0.3703 & -- & $<0.001$ & $<0.001$ & Yes \\
Ford (NM1) & RS+RLA & +0.0082 & [0.004, 0.012] & $<0.001$ & $<0.001$ & Yes \\
Ford (NM1) & RS+OLA & +0.0016 & [-0.002, 0.005] & 0.315 & 0.221 & No \\
Ford (NM1) & ES+RLA & +0.0059 & [0.003, 0.009] & $<0.001$ & $<0.001$ & Yes \\
Ford (NM1) & ES+OLA & +0.0049 & [0.002, 0.008] & 0.004 & 0.003 & Yes \\
Ford (NM1) & MV & +0.0019 & [-0.002, 0.006] & 0.331 & 0.440 & No \\
Ford (NM1) & KC+RLA & +0.0038 & [0.000, 0.007] & 0.047 & 0.084 & Yes \\
Ford (NM1) & BADGE+RLA$^\dagger$ & +0.0034 & -- & 0.026 & 0.028 & Yes \\
Music & RS+RLA & +0.0200 & [0.012, 0.028] & $<0.001$ & $<0.001$ & Yes \\
Music & RS+OLA & +0.0073 & [-0.002, 0.016] & 0.111 & 0.073 & No \\
Music & ES+RLA & +0.0077 & [-0.001, 0.016] & 0.067 & 0.109 & No \\
Music & ES+OLA & -0.0086 & [-0.015, -0.003] & 0.008 & 0.006 & Yes \\
Music & MV & +0.0942 & [0.085, 0.104] & $<0.001$ & $<0.001$ & Yes \\
Music & KC+RLA & +0.0330 & [0.025, 0.041] & $<0.001$ & $<0.001$ & Yes \\
Music & BADGE+RLA & +0.0139 & [0.005, 0.023] & 0.004 & 0.004 & Yes \\
\end{longtable}
}

\newpage

\section*{Appendix D}
\label{app.reproducibility}

\textbf{Random seeds and reproducibility.} The train/test split (below, $\texttt{random\_state=42}$) and the classifier initializations ($\texttt{random\_state=0}$) use a fixed random state, held constant across all replications and methods, so these two elements are exactly reproducible. The remaining stochastic elements of each replication, generating the labeler accuracy pool, sampling the initial labeled set, selecting query points under random sampling, assigning the specific incorrect class when a label is corrupted, resolving ties in majority voting, and the $k$-means++ seeding used by BADGE+RLA, draw from the numpy and Python standard-library random number generators without a fixed seed, and are therefore not deterministically reproducible across separate executions of the code. We report this explicitly for transparency rather than describe the pipeline as more reproducible than it is.

This does not affect the validity of the paired significance tests in Appendix~C: pairing there relies on the seven core methods (RS+RLA, RS+OLA, ES+RLA, ES+OLA, OLAS, MV, KC+RLA) being computed inside the same per-replication function call within a single execution, so that a given replication's initial labeled set, labeler accuracy pool, and train/test split are identical across those seven methods, regardless of whether that execution could be exactly reproduced later. BADGE+RLA is run as a separate script execution and does not share this per-replication state with the other methods, so it is compared using unpaired tests instead (see Appendix~C).

\textbf{Train/test split.} For the five UCI/Covertype datasets, an 80/20 train/test split is used (\texttt{shuffle=True}, \texttt{random\_state=42}), held constant across all replications and methods.

\textbf{Initialization.} For the five UCI/Covertype datasets, the initial labeled set is formed by stratified sampling of three instances per class (fewer if a class has fewer than three available instances). For the Ford case study, the initial labeled set is a random 16\% of the training set (Section~\ref{sec.casestudy}).

\textbf{Preprocessing.} Features are standardized (zero mean, unit variance) for use in KC+RLA's distance-based sample selection; no additional preprocessing is applied for the other methods or classifiers.

\textbf{F1 averaging.} Binary F1 is reported for the four two-class UCI datasets (Statlog Heart, Ionosphere, Connectionist Bench Sonar, Spambase); macro-averaged F1 is reported for Covertype, the Music dataset, and the Ford case study, all of which have more than two classes.

\textbf{Classifier hyperparameters.} A random forest classifier (\texttt{n\_estimators=50}, \texttt{random\_state=0}, \texttt{n\_jobs=-1}; otherwise \emph{scikit-learn} default settings) is used for all methods on the four smaller UCI datasets, the Music dataset, and the Ford case study. An extra-trees classifier (\texttt{n\_estimators=20}, \texttt{max\_depth=10}) is used in place of random forest for Covertype, to manage computational cost at that scale (Section~\ref{sec.benchmark}).

\textbf{Hardware and software.} All experiments were run on a machine with an Apple Silicon (arm64) processor, 10 CPU cores, and 16 GB RAM, using Python 3.9.6 with numpy 2.0.2, scipy 1.13.1, pandas 2.3.3, scikit-learn 1.6.1, and joblib 1.5.3 (used to parallelize replications across CPU cores).

\textbf{Code and data availability.} Code implementing OLAS and all baseline methods, the noise-estimation pipeline, and the significance tests reported in this paper is publicly available at
\href{https://github.com/AhadiPouya/OLAS-Optimal-Labeler-Assignment-and-Sampling-for-Active-Learning}{https://github.com/AhadiPouya/OLAS-Optimal-Labeler-Assignment-and-Sampling-for-Active-Learning}.
The five UCI/Covertype datasets and the Music Genre Classification dataset are all publicly available from their original sources, cited in Section~\ref{sec.benchmark} and Section~\ref{sec.music} respectively; the repository does not redistribute them. The Ford Motor Company warranty claim dataset used in Section~\ref{sec.casestudy} is proprietary and is not publicly available, consistent with the confidentiality agreement under which it was obtained; the code implementing the same OLAS methodology for that case study is likewise withheld, since it depends on that proprietary data's structure. A full mapping from each reported result to the specific code file and dataset that produce it, along with a detailed installation guide, is provided in the accompanying IJDS Data and Code Reproducibility Report submitted with this manuscript.

\end{document}